\documentclass[11pt,a4paper]{article}
\usepackage[margin=1in]{geometry} 
\usepackage{cite}
\usepackage{amsmath,amssymb,amsfonts}
\usepackage{algpseudocode}
\usepackage{graphicx}
\usepackage{textcomp}
\usepackage{float}
\usepackage{xcolor}
\usepackage{booktabs}
\usepackage{multirow}
\newtheorem{theorem}{\bf Theorem}
\usepackage{array}
\usepackage{xcolor} 
\usepackage{threeparttable}
\usepackage{subcaption}
\usepackage{tabularx}
\usepackage{adjustbox}
\usepackage{boldline}
\usepackage{xspace}
\usepackage{listings}
\usepackage{multicol} 
\definecolor{Pine}{RGB}{0,139,114}
\definecolor{Brick}{RGB}{182,50,28}
\definecolor{Cerulean}{RGB}{0,162,227}
\definecolor{dodgerblue}{RGB}{30,144,255}
\usepackage{hyperref}
\usepackage{changepage}
\usepackage{subcaption}
\usepackage{algorithm}
\usepackage{booktabs}
\usepackage{tabularx}
\usepackage{multirow}
\usepackage{verbatim}
\usepackage{wrapfig}
\usepackage{soul}
\usepackage{caption}

\newcommand{\boldres}[1]{{\textbf{#1}}}
\newcommand{\secondres}[1]{\underline{#1}}

\newcommand{\methodn}{\texttt{FreIE}}
\newcommand{\method}{\textsc{\methodn}\xspace}

\title{FreIE: Low-Frequency Spectral Bias in Neural Networks for Time-Series Tasks}

\author{
    Jialong Sun$^{1}$,
    Xinpeng Ling$^{2}$,
    Jiaxuan Zou$^{3}$,
    Jiawen Kang$^{4}$,
    and Kejia Zhang$^{5,}$\thanks{Corresponding author: zhangkejia@hlju.edu.cn}
}

\date{} 

\begin{document}
\maketitle

\begin{center}
\small 
$^1$School of Mathematical Science, Heilongjiang University, Harbin, China \\
\texttt{20212644@s.hlju.edu.cn} \\[1ex] 

$^2$Software Engineering Institute, East China Normal University, Shanghai, China \\
\texttt{xpling@stu.ecnu.edu.cn} \\[1ex]

$^3$Mathematics and Statistics, Xi'an Jiaotong University, Xi'an, China \\
\texttt{jiaxuanzou@stu.xjtu.edu.cn} \\[1ex]

$^4$School of Automation, Guangdong University of Technology, Guangzhou, China \\
\texttt{kavinkang@gdut.edu.cn} \\[1ex]

$^5$School of Computer Science and Big Data (School of Cybersecurity), Heilongjiang University, Harbin, China \\

\end{center}

\begin{abstract}
The inherent autocorrelation of time series data presents an ongoing challenge to multivariate time series prediction. Recently, a widely adopted approach has been the incorporation of frequency domain information to assist in long-term prediction tasks. Many researchers have independently observed the spectral bias phenomenon in neural networks, where models tend to fit low-frequency signals before high-frequency ones. However, these observations have often been attributed to the specific architectures designed by the researchers, rather than recognizing the phenomenon as a universal characteristic across models. To unify the understanding of the spectral bias phenomenon in long-term time series prediction, we conducted extensive empirical experiments to measure spectral bias in existing mainstream models. Our findings reveal that virtually all models exhibit this phenomenon. To mitigate the impact of spectral bias, we propose the FreLE (Frequency Loss Enhancement) algorithm, which enhances model generalization through both explicit and implicit frequency regularization. This is a plug-and-play model loss function unit. A large number of experiments have proven the superior performance of FreLE. Code is available at https://github.com/Chenxing-Xuan/FreLE.
\vspace{1em}

\noindent\textbf{Keywords:} Time series forecasting, Fourier transform, Implicit Regularization.
\end{abstract}

\section{Introduction}
Time series data consists of numerical values associated with time. Long-term time series prediction is crucial across various domains, including weather forecasting and intelligent manufacturing~\cite{weather1,IOT2}. However, due to the inherent complexity of time series data, existing deep learning approaches that directly predict time-domain data often yield suboptimal performance. In recent years, a promising approach has emerged that leverages frequency-domain information to improve prediction accuracy.

Modeling long-term time series prediction using quasi-periodic dynamical systems reveals that both linear and nonlinear time-domain prediction optimization objectives are highly non-convex. However, by mapping the optimization objective to the frequency domain, the global optimal solution of the error surface can be efficiently computed using the Koopman-FFT method~\cite{fourier1}. This theoretical foundation has significantly inspired researchers to incorporate frequency-domain information into long-term time series prediction. Building on Koopman's work, a method has been proposed that transforms frequency-domain information into 2D, converting frequency sequence data into frequency image data. This method employs 2D kernel modeling to capture implicit frequency relationships between different sequences, thereby enhancing time-domain learning performance~\cite{Timenet}. Additionally, given the complex information resulting from frequency-domain transformations, complex-valued neural networks can be employed to achieve efficient long-term time series prediction with a reduced number of parameters~\cite{FITS}. Recent studies have also provided both theoretical proof and empirical analysis demonstrating that using frequency-domain loss functions can decouple the complexity of time series~\cite{FreDF}, further improving model performance in long-term time series prediction.

However, as the saying goes, "there is no free lunch." While frequency-domain information offers researchers a potentially limitless framework for machine learning, it also presents inevitable challenges, particularly concerning the "selection of spectral information." After decomposing a signal into its spectral components, determining how to effectively utilize both low-frequency and high-frequency information within a machine learning framework has become a central area of investigation. Low-frequency signals represent stable events with higher intensity over time but fail to capture the variability of these events. In contrast, high-frequency signals reflect more volatile and trend-based events over time but are highly susceptible to noise interference. The question remains: how should these signals be leveraged in models? Based on the Johnson-Lindenstrauss Lemma, one approach employs a random dimensionality reduction method that selectively chooses specific frequency signal features for auxiliary prediction, effectively mitigating noise interference in high-frequency features~\cite{Fedformer}. Another method, grounded in the Parseval Theorem, proposes a multilayer perceptron (MLP) model architecture that applies equal signal strength in both the time and frequency domains to jointly learn time-domain signal features~\cite{FreTS}. While these approaches have significantly improved long-term time series prediction performance, they have yet to fully address the original question. How should we truly understand the role of spectral information in time series prediction?

Interestingly, when researchers investigate the role of spectral information in time series prediction, they often reach the same conclusion: in implicit neural representations (INR) networks, a tendency toward simple solutions is observed during the reconstruction process, with most solutions being linear combinations of low-frequency signals~\cite{TSINR}. In studies of Transformer attention mechanisms, researchers have found that, during prediction, the Transformer architecture first learns low-frequency signal features before progressing to high-frequency signal features~\cite{Fred}. This learning sequence is believed to be influenced by the attention mechanism's inherent bias toward low-frequency signals. While these findings provide in-depth insights into the mechanisms of frequency learning, the researchers unanimously agree that this frequency preference phenomenon is an intrinsic characteristic of specific models. In the following, we will refer to this as the "spectral bias phenomenon" and conduct a comprehensive investigation.

Some researchers have examined the "spectral bias phenomenon" from the perspective of numerical solutions to partial differential equations in neural networks. When solving the Poisson-Boltzmann equation, decomposing the loss function into low-frequency and high-frequency components can significantly enhance numerical stability. Colleagues have conducted extensive experiments to verify the existence of the "spectral bias phenomenon" in two-layer deep neural networks (2-DNNs) and provided theoretical proof of its presence in two-layer infinitely wide DNNs~\cite{Freq2}. Furthermore, a variational dynamics theory based on linear assumptions confirmed the "spectral bias phenomenon" in existing neural networks. The theory proposed that this phenomenon primarily depends on the nonlinear transformation of the activation function and recommended using the Ricker activation function to mitigate it~\cite{Freq3, Freq4}. However, the question remains: \emph{\textbf{Can this approach be extended to time series prediction tasks? Is there a simpler method for understanding and addressing the "spectral bias phenomenon"?}} This remains an unresolved issue in the field of time series prediction.

This work aims to investigate the existence of the spectral bias phenomenon in neural networks with various architectures and to improve time series prediction performance by addressing this phenomenon. We introduce FreLE (Frequency Loss Enhancement), an adaptive frequency enhancement algorithm designed to mitigate the spectral bias phenomenon observed in neural networks during time series prediction tasks. To validate the effectiveness of our method, we conducted extensive preliminary experiments and compared it with existing machine learning methods, highlighting the similarities and differences between various approaches. The main contributions of this paper are summarized as follows:
\begin{itemize}
    \item \textbf{Theoretical Research:} Building on the existing 2-DNN spectral bias dynamics theory, we conduct extensive empirical research on existing temporal neural networks. Our findings confirm that various neural network architectures exhibit the spectral bias phenomenon. 
    \item \textbf{Algorithm Design:} The FreLE algorithm we propose consists of two key components: frequency explicit regularization and frequency implicit regularization. These components are designed to perform two tasks—denoising and balancing signals of different frequencies. The roles and irreplaceability of these components are further analyzed through ablation experiments.
    \item \textbf{Experimental Effect:} We conducted extensive experiments to validate the effectiveness of FreLE, which achieved first place 38 times and second place 18 times across seven real-world datasets, demonstrating its theoretical superiority.
\end{itemize}

\begin{figure}[htbp]
    \centering
    \begin{subfigure}[b]{0.45\textwidth}
        \centering
        \includegraphics[width=1\textwidth]{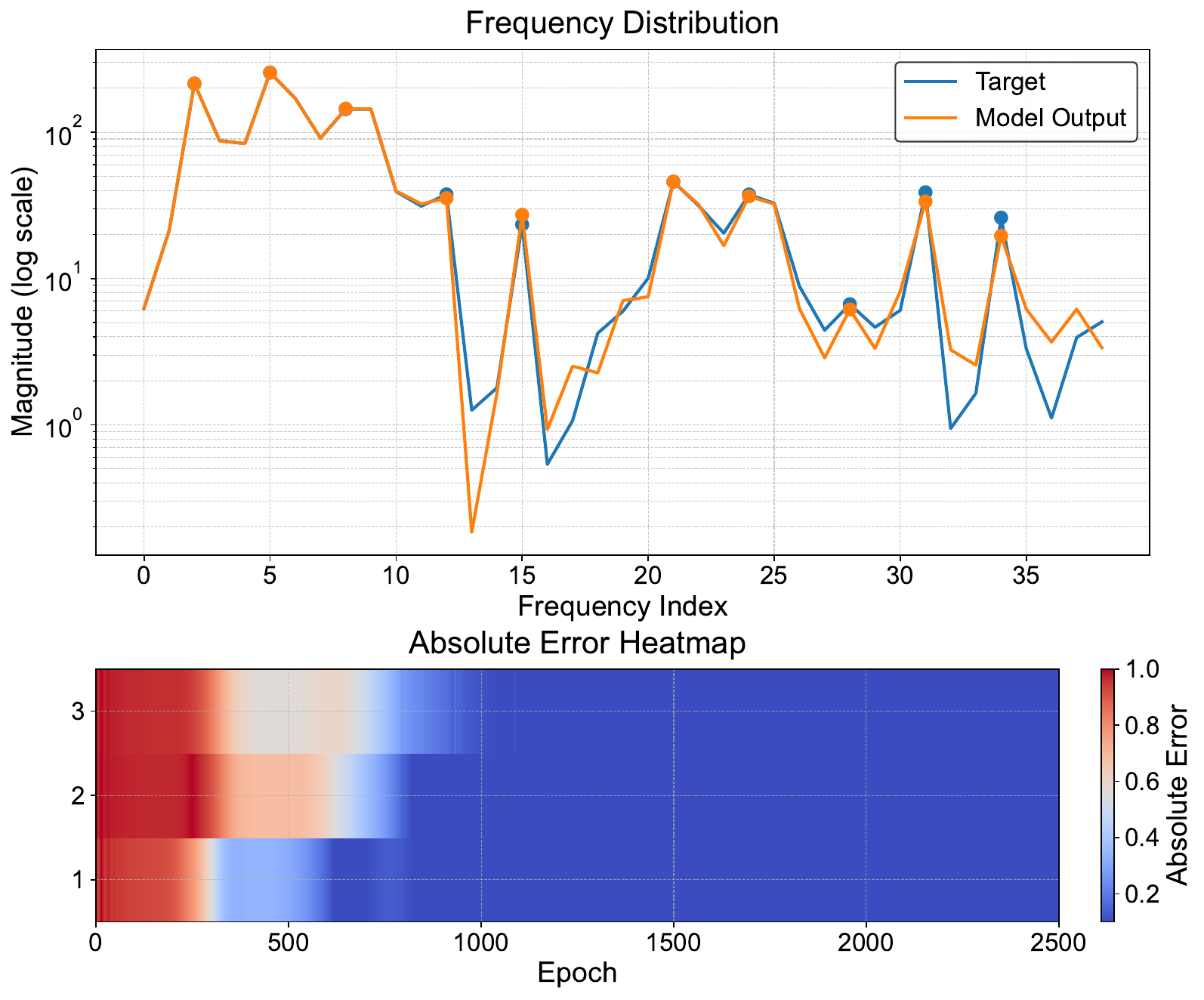}
        \caption{The spectral loss graph for synthetic data $y = \sin x + \sin 2x +  \sin 3x $}
        \label{fig:subfig1}
    \end{subfigure}
    \hfill
    \begin{subfigure}[b]{0.45\textwidth}
        \centering
        \includegraphics[width=1\textwidth]{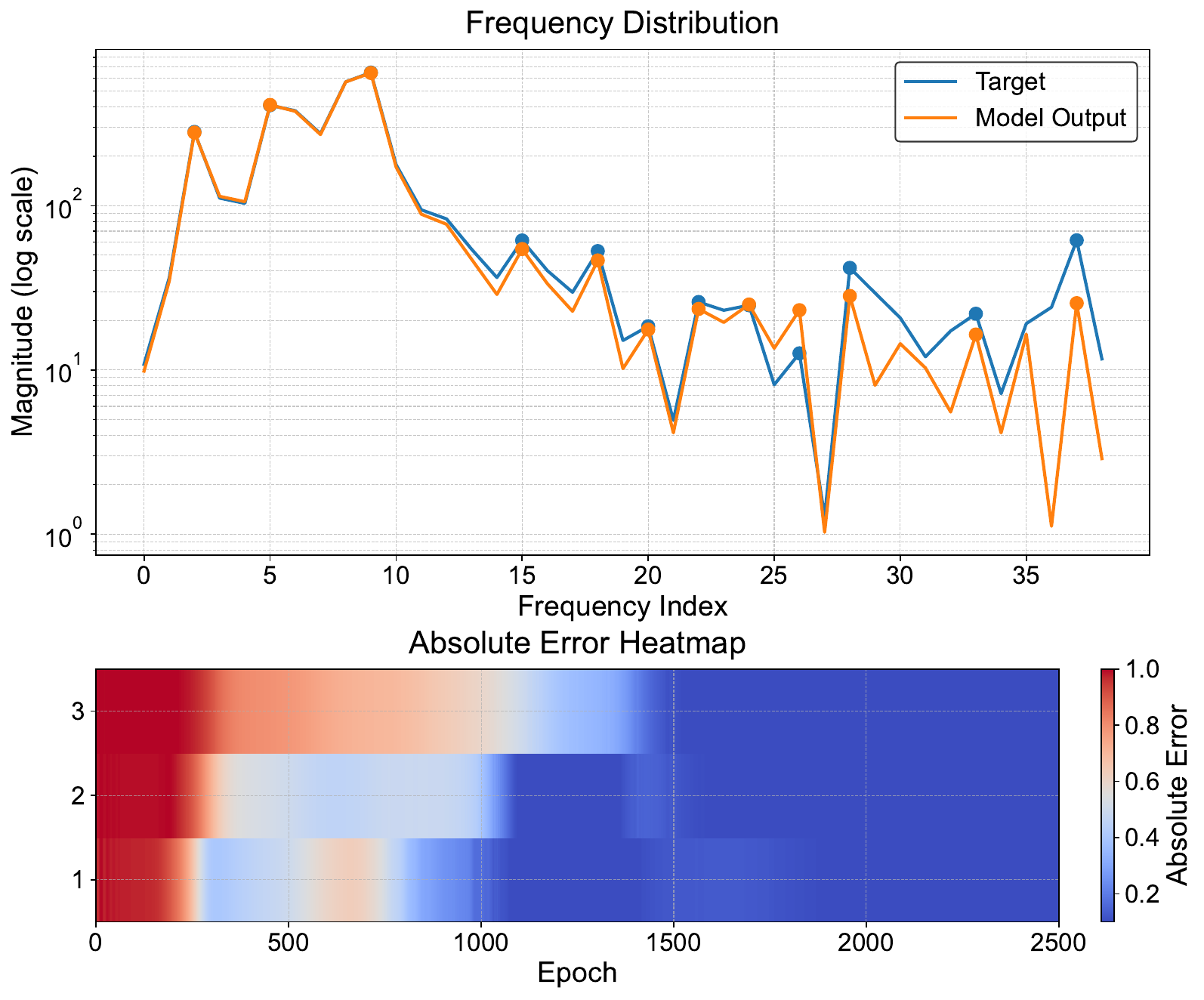}
        \caption{The spectral loss graph for synthetic data$y = \sin x + 2\sin 2x + 3 \sin 3x $}
        \label{fig:subfig2}
    \end{subfigure}
    \caption{The spectral loss graph for a 2-DNN across different synthetic datasets. The line graph represents represents the frequency comparison between the original data and the output data in the final iteration, and the heatmap illustrates the decrease in the RMSE loss metric as the iterations progress, showing how the three primary frequencies change with the number of iterations}
    \label{fig:FLoss-1}
\end{figure}
\begin{figure}[htbp]
    \centering
    \begin{subfigure}[b]{0.45\textwidth}
        \centering
        \includegraphics[width=1\textwidth]{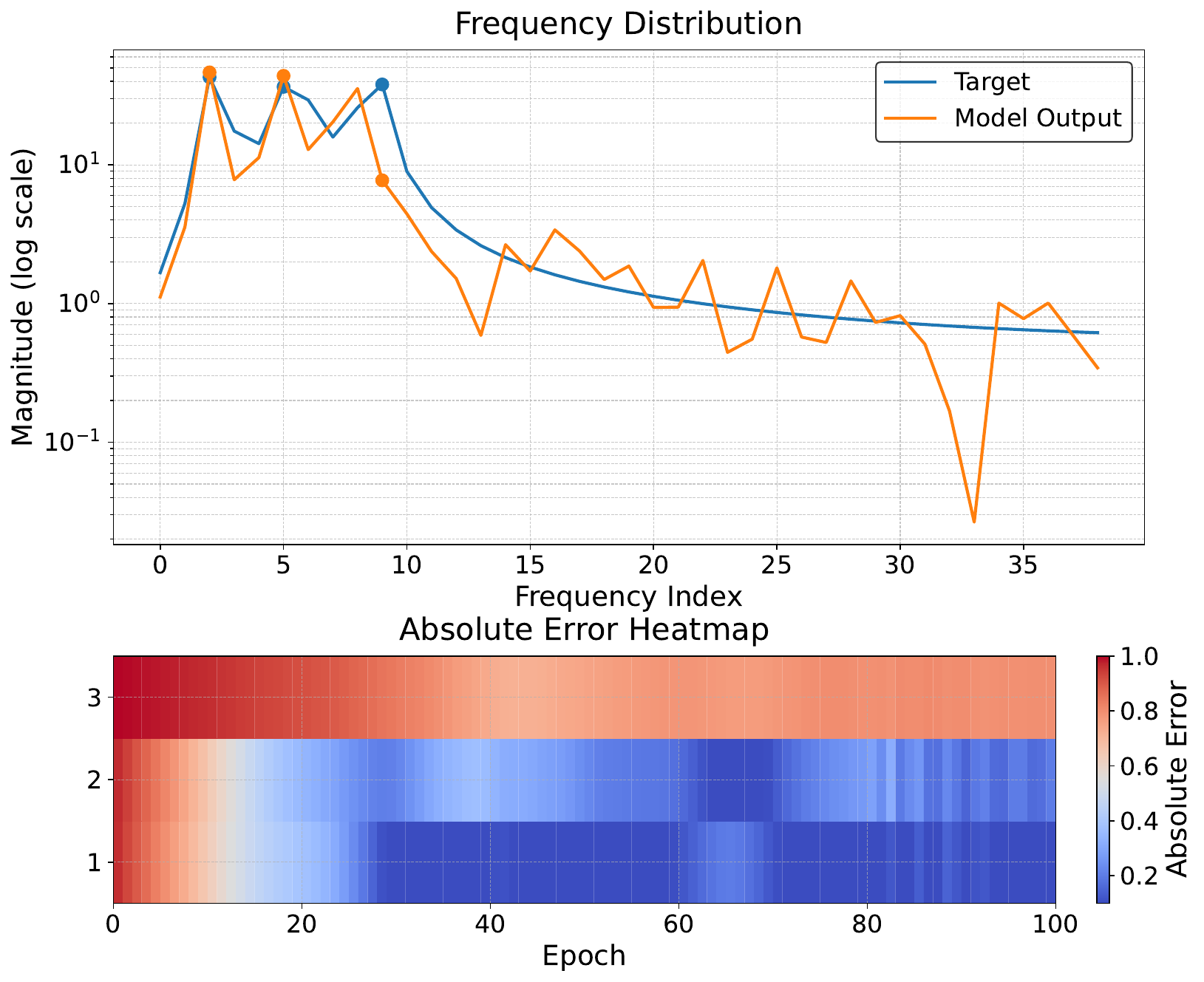}
        \caption{The spectral loss graph based on $\sigma_{\mathrm{relu}}$}
        \label{fig:subfig1}
    \end{subfigure}
    \hfill
    \begin{subfigure}[b]{0.45\textwidth}
        \centering
        \includegraphics[width=1\textwidth]{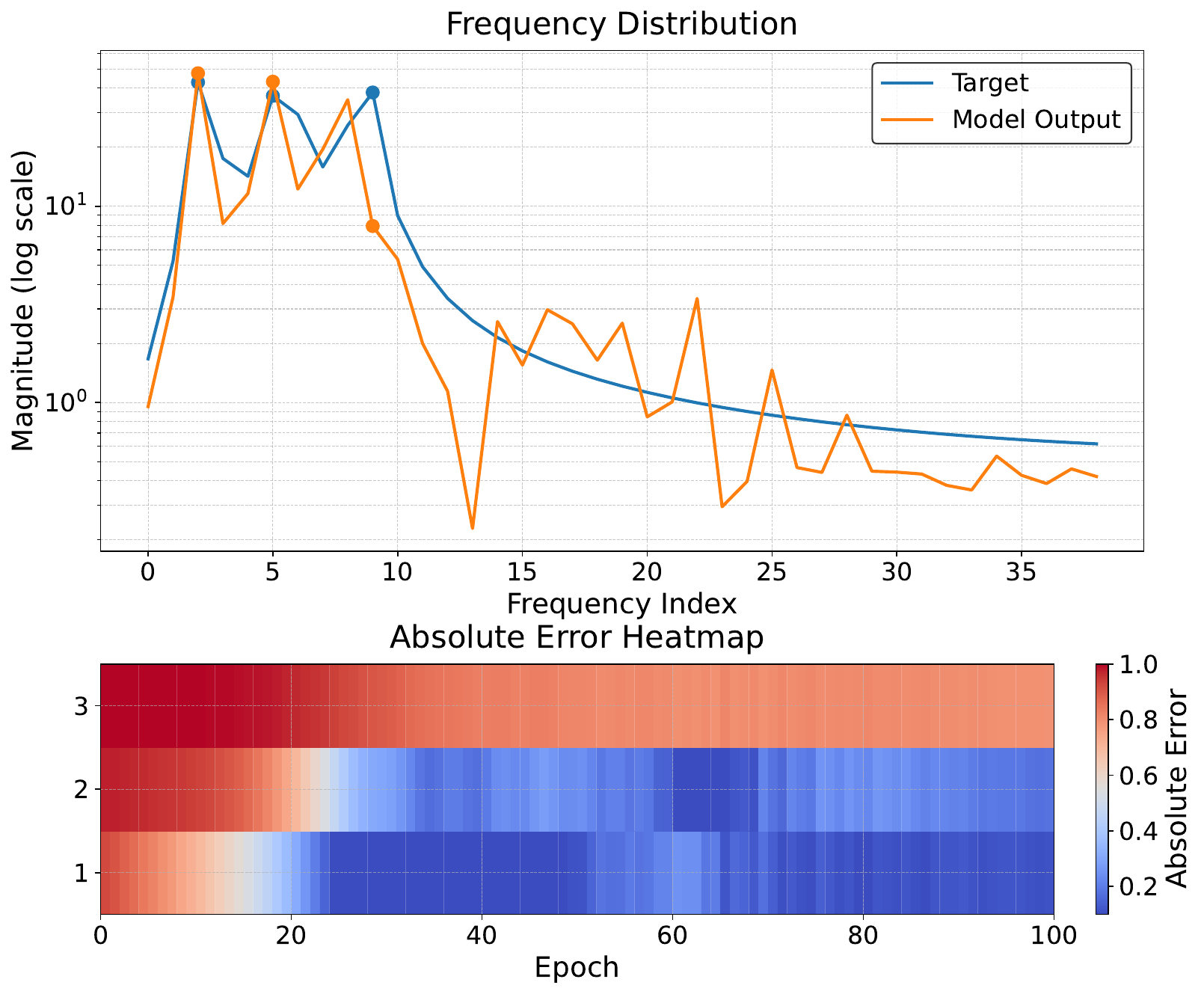}
        \caption{The spectral loss graph based on $\sigma_{\mathrm{tanh}}$}
        \label{fig:subfig2}
    \end{subfigure}
    \caption{The spectral loss graph for LSTM across different $\sigma$. }
    \label{fig:FLoss-21}
\end{figure}
\section{Preliminary Analysis}
We investigate the spectral bias phenomenon in neural networks through three different experiments: 1) examining the spectral bias phenomenon in simple time series models using synthetic datasets; 2) exploring the spectral bias phenomenon in various classical models on real-world datasets; 3) evaluating the effectiveness of existing dynamic theories in mitigating the spectral bias phenomenon. The symbols and formulas associated with the spectral dynamics theory for neural networks are introduced in Sec. 2.1. The experimental analysis of synthetic datasets is presented in Sec. 2.2. The analysis of real-world datasets is discussed in Sec. 2.3. The experimental improvements based on dynamic theory are outlined in Sec. 2.4.

\subsection{Spectral Dynamics in Neural Networks}
This section will explore three key aspects of the study of spectral dynamics in neural networks: spectral visualization, spectral dynamics hypotheses, and the formulation of spectral dynamics equations under different activation functions.

\subsubsection{Spectral Visualization}
In machine learning theory, a phenomenon related to spectral bias that has garnered the attention of mathematicians: when using synthetic data formed by the sum of multiple sine signals (e.g., \( y = \sin x + 2\sin 2x + 3 \sin 3x \)) for deep learning training, it is often observed that as the number of iterations increases, the low-frequency signals converge rapidly, while the high-frequency signals converge more slowly~\cite{Freq2}. Before extending this issue to the time series domain, we replicate the spectral bias phenomenon in 2-DNNs to facilitate further in-depth discussion. The visual representation is shown in Fig.~\ref{fig:FLoss-1}.

\subsubsection{Spectral Dynamics Formula}

For the loss function of a two-layer wide neural network, its Fourier expansion can be expressed, and the relationship between the derivative of the expanded loss function \( \mathcal{F} \) with respect to time \( t \) can be described as follows~\cite{NTK1, NTK2, NTK3, NTK4, Freq3, Freq4}:
\begin{equation}
\begin{split}
           &\partial_t \mathcal{F}[u](\xi, t) = -\mathcal{L}[\mathcal{F}[u_\rho]],\\
   &\mathcal{L}[\mathcal{F}[u_\rho]] \approx \frac{\Gamma^*(d/2)}{\|\xi\|^{d-1}} E \left[ a^{-1}(0) \mathcal{F}[\mathbf{K}] H(\xi) \right] \mathcal{F}[u_\rho](\xi),\\
     & \Gamma^*(d/2)=\frac{\Gamma(d/2)}{2\sqrt{2}\pi (d+1)/2 \sigma},\\
     &  H(\xi)=-\frac{\|\xi\|}{a(0)},
\end{split}
\end{equation}
where, $F[u]$ and $F[u_P]$ represent the loss functions under different sampling densities. $F[u_P]$ describes the samples obtained from the current iteration of training, where the loss function is influenced solely by the batch size of data in a single iteration. $\xi$ represents the frequency of the current signal, $a(0)$ is the randomly initialized weight of the neural network's linear layer, and $b(0)$ is the randomly initialized weight of the neural network's activation function. $\sigma$ denotes the activation function, and $\mathbf{K}(x) \triangleq (\sigma(x), b \sigma'(x))'$. Therefore, the form of the Linear Frequency Principle (LFP) derived above is closely related to the choice of the activation function $\sigma$, and the general form of the Fourier expansion of the loss function can be obtained as follows:
\begin{equation}
     \partial_t \mathcal{F}[u](\xi, t)=(\gamma_{\sigma}(\xi))^2[\mathcal{F}[u_\rho]] 
\end{equation}
where, $\gamma_{\sigma}(\xi)$ is the frequency decay function obtained for different activation functions. Theorem 2 emphasizes the expression that \textbf{after performing a Fourier transform on the loss function, different frequencies follow different decay schemes in the gradient expression.} This decay scheme is often related to the choice of activation function. More specifically, the expressions for $\gamma_\sigma$ in the ReLU and Tanh activation functions are shown in Theorem ~\ref{The:1}.
\begin{theorem}
    (Decay functions of different activation functions)\textbf{.} The \( \gamma_{relu} \) of the ReLU activation function can be expressed as:
    \begin{equation}
(\gamma^2_{\mathrm{relu}}(\xi)) = \mathbb{E} \left[ \frac{a(0)^3}{16\pi^4 \lVert \xi \rVert^{d+3}} + \frac{b(0)^2 a(0)}{4\pi^2 \lVert \xi \rVert^{d+1}} \right].
\end{equation}

The \( \gamma_{\mathrm{tanh}} \) of the tanh activation function can be expressed as:
\begin{equation} 
\begin{aligned}
(\gamma^2_{\mathrm{tanh}}(\xi)) =& \frac{1}{\lVert \xi \rVert^{d-1}} \mathbb{E}_{a,r} \left[ \frac{\pi^2}{r} \text{csch}^2 \left( \frac{\pi \lVert \xi \rVert}{r} \right)\right. \\
&+ \left.\frac{4\pi^4 a^2 \lVert \xi \rVert^2}{r^3} \text{csch}^2 \left( \frac{\pi \lVert \xi \rVert}{r} \right) \right].
\end{aligned}
\end{equation}\label{The:1}
\end{theorem}

Theorem ~\ref{The:1} shows that the spectral bias decays according to the power of the frequency of the spectral signal. As the frequency increases, the gradient of the loss function rapidly decays to zero. This represents a classical dynamical theoretical analysis in machine learning~\cite{Freq3, Freq4, Freq5}. However, this theoretical result also raises a new issue:\emph{\textbf{Q: Some classical time series models, such as RLinear, DLinear, and FITS\cite{RLinear, DLinear, FITS}, do not incorporate activation functions during the prediction process. Therefore, is the frequency preference principle widely observed in time series models? Can it be improved by introducing or modifying the activation function?}}

\textbf{A:}In the subsequent experiments of  Secs. 2.2-2,5, more extensive empirical tests will be conducted to demonstrate that the spectral bias phenomenon in time series tasks cannot be solely attributed to the effects of activation functions. The spectral bias phenomenon is widely observed in both linear and nonlinear models. Moreover, alleviating spectral bias in 2-DNNs by modifying the activation function did not yield favorable results in the time series domain. Instead, the activation function's hyperparameters significantly impacted the convergence speed and final performance.

\subsection{LSTM Experiment on Synthetic Datasets}
Based on the experiments in Sec. 2.1.3, the 2-DNN was replaced with an LSTM neural network for training, with the activation functions being ReLU and Tanh. In a large number of experiments, significant spectral phenomena were still observed. Some experimental results are shown in Fig. ~\ref{fig:FLoss-21}. 

\subsection{Experiments on Real-World Datasets}
We have compiled a list of classic time series models from recent years and categorized them based on their structure into two main categories: MLP models and Transformer models. These categories are divided into two subcategories: whether frequency domain data was incorporated during the training process. The models discussed are summarized as follows: 1) MLP models without frequency domain data: DLinear \cite{DLinear}, Tide \cite{Tide}; 2) MLP models with frequency domain data: TimesNet \cite{Timenet}, FreTS \cite{FreTS}, FreDF \cite{FreDF}, FITS \cite{FITS}; 3) Transformer models without frequency domain data: Autoformer \cite{Auto}, CrossFormer \cite{Cross}; 4) Transformer models with frequency domain data: FEDformer \cite{Fedformer}. 

In addition, we define the top 10\% of the Fourier transform frequency results as low-frequency signals, the 10\%-50\% range as mid-frequency signals, and the 50\%-100\% range as high-frequency signals. We also introduce the concept of global signals, referring to the entire signal obtained after the Fourier transform. The results of the spectral bias phenomenon calculated on the ETTH2 dataset are presented in Table \ref{tab:F}. It can be observed that most models with strong predictive performance are based on frequency-domain information. Meanwhile, the strict spectral bias phenomenon appears in all models except TimesNet, which exhibits convergence characteristics in the high-frequency range. This is due to the 2D-FFT performed by TimesNet, which disrupts frequency information and causes overfitting of high-frequency signals. The low-frequency and mid-frequency signals, both of equal importance, are not well fitted, resulting in the model having the best frequency fitting but weaker performance than other models.

\begin{table}[ht] 
\caption{For the long-term forecasting task on the ETTH2 dataset, LIL is configured with a past sequence length of 36, while other settings are set to 96. Models marked with an asterisk * use frequency information to assist in prediction. LF, MF, HF, and GF represent low-frequency, mid-frequency, high-frequency, and global frequency, respectively. The evaluation metric used is RMSE. The \textbf{best} results are highlighted. }
  \centering
  {\scriptsize

    \begin{tabular}{l|*{4}{c}|*{2}{c}}
    \toprule
    Domain & \multicolumn{4}{c|}{Frequency domain indicators} & \multicolumn{2}{c}{Time domain indicators} \\
    \midrule
    Metrics & LF & MF & HF & GF & MAE & MSE  \\
    \midrule
        Delinear & 0.5635 & 0.6261 & 1.7671 & 1.0598 & 0.3963 & 0.3425  \\
        Tide & 0.3951 & 0.8379 & 0.8641 & 0.6643 & 0.3384 & 0.2894  \\
    \midrule
        TimesNet$^*$ & 0.3994 & 0.6707 & \textbf{0.0481} & \textbf{0.2453} & 0.3640 & 0.3198  \\
        FreTS$^*$ & 0.7969 & \textbf{0.2338} & 1.5131 & 0.9283 & 0.4043 & 0.3511  \\
        FreDF$^*$ & 0.2963 & 0.9051 & 1.4407 & 1.0087 & 0.3438 & 0.2940  \\
        FITS$^*$ & \textbf{0.1476} & 0.7151 & 1.0750 & 0.8291 & \textbf{0.3367} & \textbf{0.2718}  \\
     \midrule
      Crossformer & 0.5496 & 0.9393 & 2.1264 & 1.3737 & 0.5925 & 0.6985 \\
        Autoformer & 0.2551 & 0.6650 & 1.6012 & 0.8788 & 0.4230 & 0.3972  \\
        Transformer & 1.8012 & 2.3223 & 2.9539 & 1.9031 & 1.1441 & 2.0782  \\
    \midrule
        Fedformer$^*$ & 0.4671 & 1.3907 & 1.4540 & 0.8367 & 0.3912 & 0.3470 \\
    \bottomrule
    \end{tabular}}
    \label{tab:F}
\end{table}
\begin{figure}[htbp]
    \centering
    \begin{subfigure}[b]{0.48\textwidth}
        \centering
        \includegraphics[width=1\textwidth]{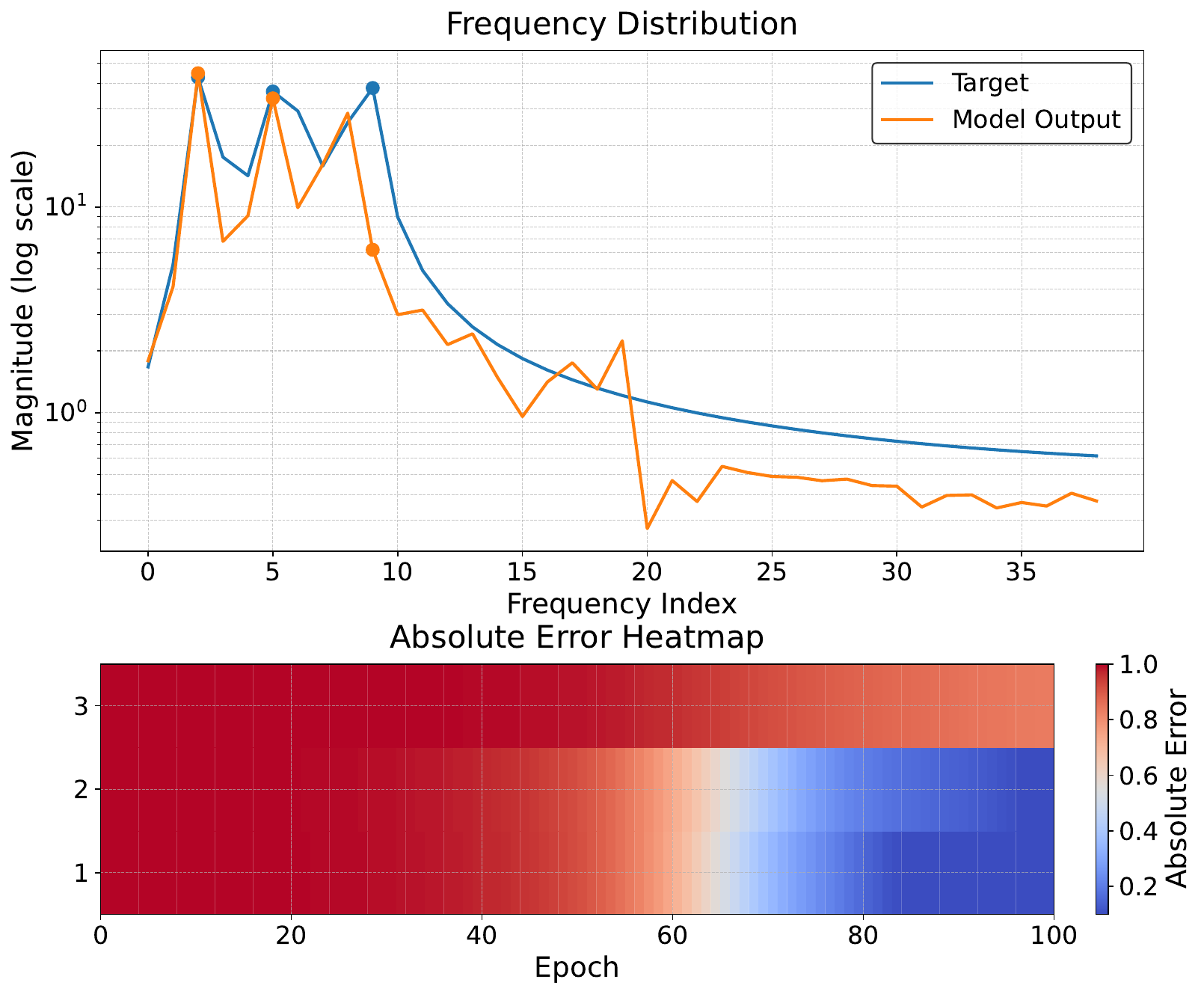}
        \caption{The spectral loss graph for synthetic data $y = \sin x + \sin 2x +  \sin 3x $}
        \label{fig:subfig1}
    \end{subfigure}
    \hfill
    \begin{subfigure}[b]{0.48\textwidth}
        \centering
        \includegraphics[width=1\textwidth]{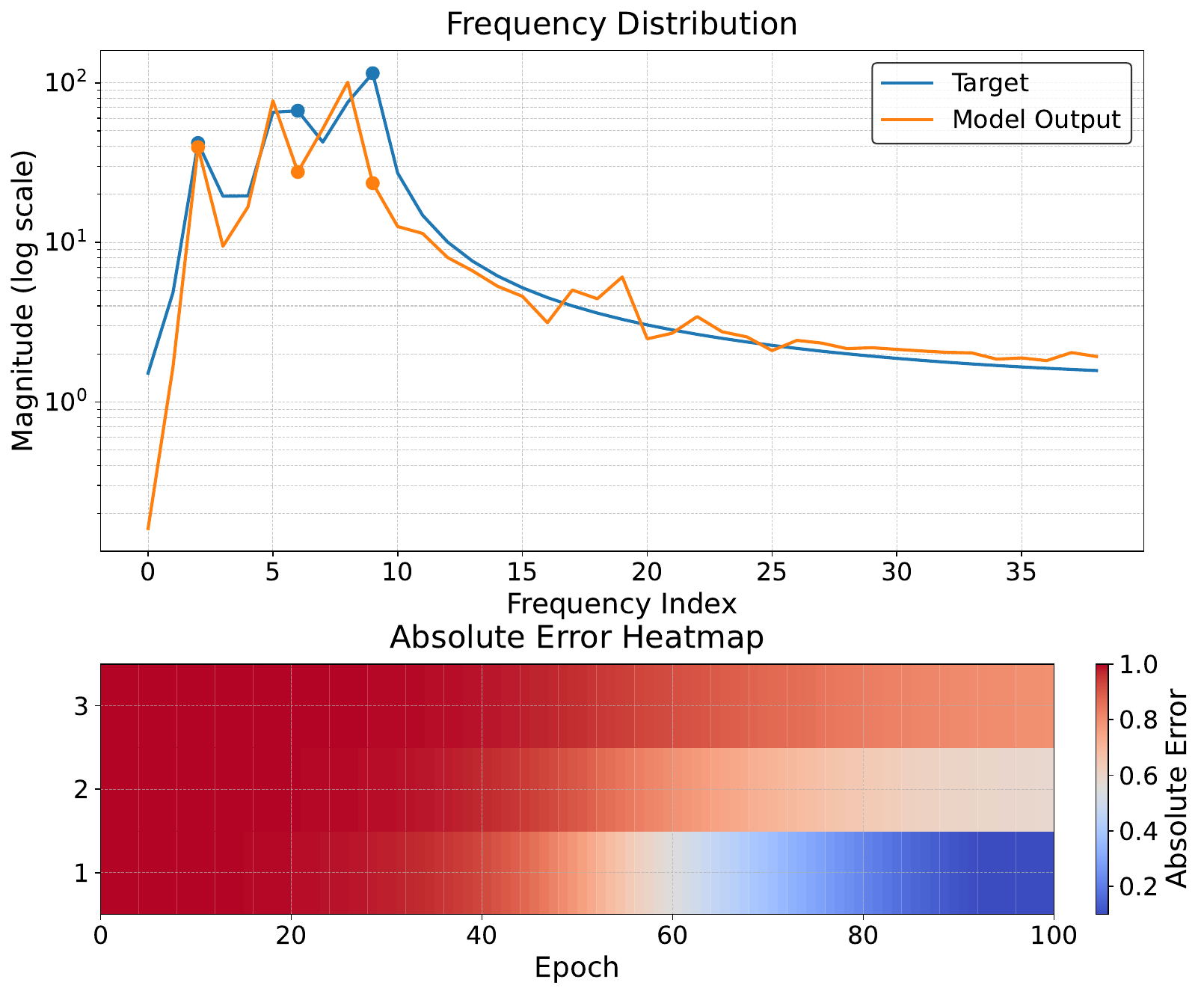}
        \caption{The spectral loss graph for synthetic data $y = \sin x + 2\sin 2x +  3\sin 3x $}
        \label{fig:subfig2}
    \end{subfigure}
    \caption{Under different synthetic datasets, frequency amplitude loss diagram of $\sigma_{\mathrm{ricker}}$ with $a=1$. }
    \label{fig:r-2}
\end{figure}
\subsection{Neural Network Optimization Based on Dynamic Theory}
In the research by Xv et al.~\cite{Freq3}, it is stated that the impact of spectral bias can be mitigated by disrupting the monotonicity of activation functions. The classical wavelet transform function, Ricker~\cite{Ricker}, demonstrates excellent expressive performance within a 2-DNN, with its mathematical expression given by:
\begin{equation}
    \sigma_{ricker}=\frac{\pi^{1/4}}{15a}  \left( 1 - \left( \frac{x}{a} \right)^2 \right) \exp \left( -\left( \frac{x}{\sqrt{2}a} \right)^2 \right)
\end{equation}
where $a$ is an adjustable hyperparameter.

In the empirical experiments of this section, we will investigate whether replacing the activation function in LSTM with Ricker leads to improved performance. Some experimental results are shown in Figure~\ref{fig:r-2}. It can be observed that while Ricker mitigates spectral bias in some experiments, its effect is negligible when dealing with more complex spectral phenomena, even with minor changes to the coefficients of different signal components. This suggests the need to explore an entirely new approach to address the issue of spectral bias in time series forecasting tasks.
\section{Frequency Enhancement: Methods}
\begin{figure}
    \centering
    \includegraphics[width=0.8\linewidth]{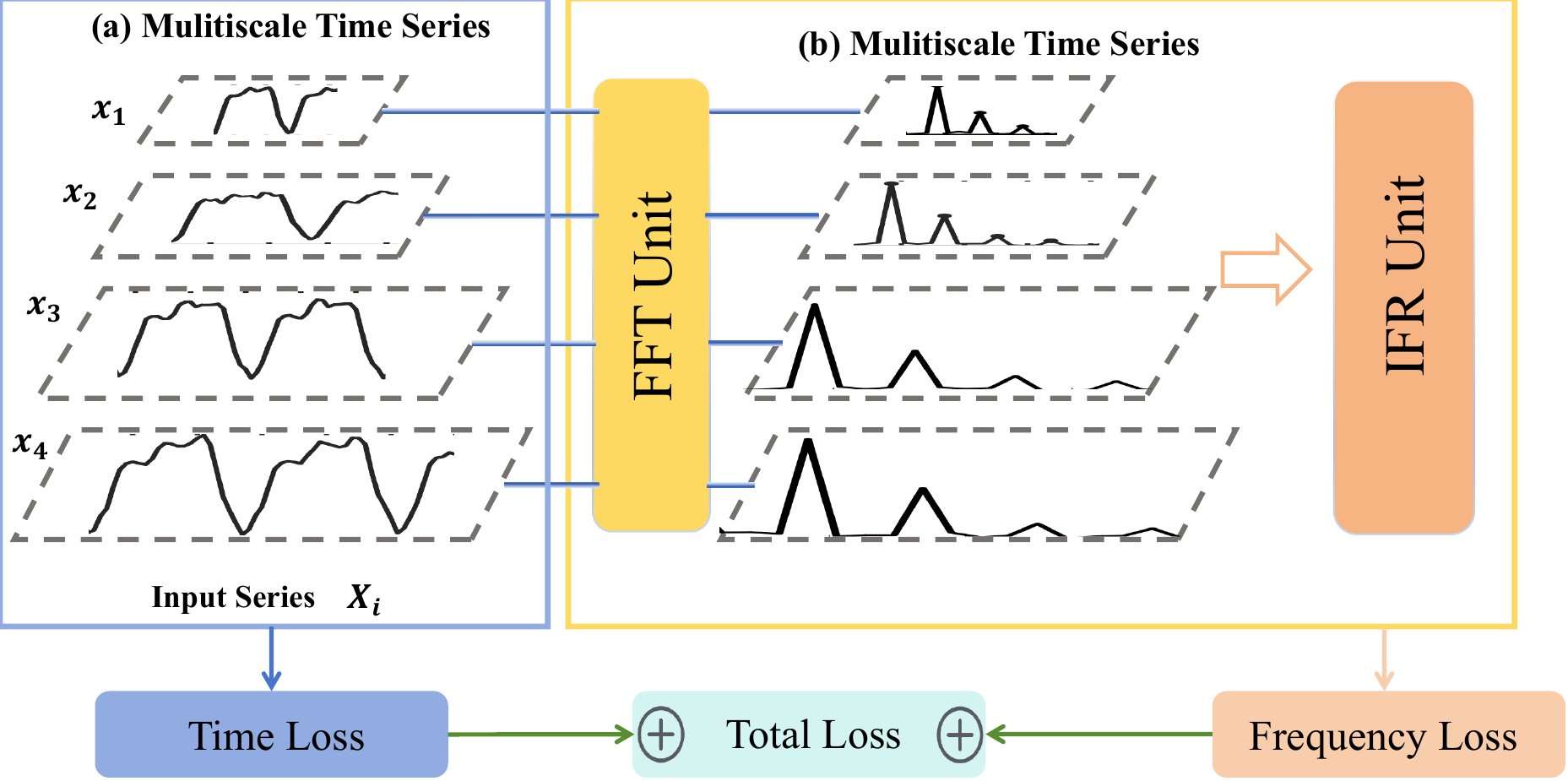}
    \caption{The framework diagram of FreLE, where IFR Unit refers to the Implicit Frequency Regularization module.}
    \label{fig:FreLE}
\end{figure}

In this section, we will elaborate on how the FreLE algorithm balances frequency information and removes noise by separately discussing its two key components: explicit frequency regularization and implicit frequency regularization. The framework diagram of FreLE is shown in Figure~\ref{fig:FreLE}.
\subsection{Explicit Frequency Regularization}
For a given time series \( X \) and its predicted value \( \hat{X} \), the time series forecasting task can be described as an optimization problem: 
\begin{equation}
\begin{aligned}
    \min_{\theta}&\quad\mathcal{L_\theta}^t\\
    \mathcal{L}_\theta^t&=\frac{1}{n}\sum_{i=1}^n\|X_i-\hat{X_i}\|
 \end{aligned}\label{eq:sun}
\end{equation}

To redefine this problem explicitly with a frequency loss function~\eqref{eq:sun}, we can incorporate it as: 
\begin{equation}
\begin{aligned}
    \min_{\theta}&\quad\delta\mathcal{L_\theta
    }^f+(1-\delta)\mathcal{L}_\theta^t\\
        \mathcal{L}^f&=\frac{1}{n}\sum^{N}_{i=1}\|\mathcal{F}(X_i)-\mathcal{F_\theta}(\hat{X}_i)\|\\
 \end{aligned}\label{eq:sun}
\end{equation}
where, $\delta$ serves as a parameter for balancing between two types of losses. An interesting research question is whether, by using explicit regularization alone, significant optimization effects can already be achieved when $\delta=1$. 
\subsection{Implicit Frequency Regularization}
The purpose of explicit frequency regularization is to incorporate frequency as a penalty term, preventing the neural network from converging too quickly after fitting the low-frequency signals. This is particularly relevant because complex neural networks (e.g., iTransformer~\cite{itrans}, DLinear~\cite{DLinear}, TimeXer~\cite{TimeXer}) typically converge on the ETTh1 dataset in an average of just eight epochs. By introducing frequency as a penalty term, the model can continue learning even after reaching its original extremum. However, simply adding explicit regularization does not effectively extend the number of training epochs. In their research, Wu et al.~\cite{Wu} thoroughly explored the enhancement of model generalization through prolonged training durations. Unlike modifying the loss function with explicit regularization, implicit regularization offers a more practical approach to improving the model's generalization capability. Therefore, this section will discuss how implicit regularization can slow down the model's learning process.

Before explaining how to achieve implicit frequency regularization, first present Theorem ~\ref{the:2}.
\begin{theorem}~\label{the:2}
    (Multi-dimensional Fourier separation theorem).A two-dimensional Fourier transform can be decomposed into two one-dimensional Fourier transforms, serving as an example of the Fourier transform applied to two-dimensional vectors. It follows:
    \begin{equation}
\begin{aligned}
&\quad  \mathcal{F}(k_x, k_y) \\&= \int_{-\infty}^{\infty} \int_{-\infty}^{\infty} f(x, y) e^{-i 2 \pi (k_x x + k_y y)} \, \mathrm{d}x \, \mathrm{d}y\\
&= \int_{-\infty}^{\infty} \left[ \left( \int_{-\infty}^{\infty} f(x, y) e^{-i 2 \pi k_x x} \, \mathrm{d}x \right) e^{-i 2 \pi k_y y} \right] \mathrm{d}y
    \end{aligned}
\end{equation}
\end{theorem}

In subsequent processing, Fourier transforms of multidimensional signals will be treated as single-channel Fourier transforms to reduce computational complexity. However, directly incorporating Fourier-transformed results into neural networks introduces significant noise interference, making denoising a crucial step. Traditional windowing methods, while commonly used, fail to achieve effective denoising. Applying windows to frequency information leads to rapid attenuation of frequencies other than the primary frequency, significantly degrades neural network performance~\cite{windows1,window}. Therefore, this section will explore a novel denoising method and integrate it into the process of implicit regularization.

The latest research proposes an adaptive frequency processing approach that normalizes the amplitude adaptively across different frequency bands~\cite{Freqgui1}. This is an effective solution for handling noise in frequency information. As is well known, in the frequency information obtained from the Fourier transform, local maximum values within a small range represent more significant extremal components. Therefore, before calculating the loss function, we progressively detect whether a frequency is a local maximum within a frequency width $d$, starting from the low frequencies. If $\xi_i$ is a local maximum of the signal amplitude, we perform a signal assignment correction on it:
\begin{equation}
    \xi^*_i=\frac{i}{\eta}\xi_i
\end{equation}
where $i$ represents the number of frequency components and $\eta$ is a dimensional balance constant, the core concept lies in adjusting the parameters of the frequency components before computing the loss function. This ensures that the frequency components do not appear as independent computational graphs during gradient computation, leading to smoother gradients. The pseudo-code for FreLE, consisting of two modules, is shown in Algorithm~\ref{Al:1}.

\begin{algorithm}
\caption{FreLE Algorithm}
\begin{algorithmic}[1]
\Require Time series data $X_i$, loss balance constant $\delta$, frequency width $d$, dimensional balance constant $\eta$.
\Ensure $\delta\mathcal{L}^f+(1-\delta)\mathcal{L}^t$
\State Frequency: $f[i] \gets \mathcal{F}(X_i)$
\State Amplitude: $A[i] \gets \|f[i]\|$  
\For{each frequency $f[i]$ in $f$}
    \If{$A[i] = \max\{A[i-\lfloor\frac{d}{2}\rfloor],\dots,A[i+\lceil\frac{d}{2}\rceil]\}$} 
        \State $f[i]\gets \frac{i}{\eta}f[i]$
    \EndIf
\EndFor
\State $\mathcal{L}^f=\text{MAE}(f[i]-\hat{f[i]})$ 
\State $\mathcal{L}^t=\text{MAE}(X_i-\hat{X_i})$
\Return $\delta\mathcal{L}^f+(1-\delta)\mathcal{L}^t$
\end{algorithmic}
\label{Al:1}
\end{algorithm}

\section{Experiment}
To verify the effectiveness of FreLE, we will examine it from the following four perspectives:
\begin{itemize}
    \item[1.]\textbf{Performance:} \emph{Does FreLE work?} In Sec. 4.2, we used classical public datasets to compare the performance metrics of FreLE with classical baselines from 2020 to 2024, demonstrating FreLE's superior performance.
    \item [2.]\textbf{Echanism:} \emph{Why does it work?} In Sec. 4.3, ablation experiments are conducted on the two existing modules separately, and the frequency signal processing method proposed in the 2024 paper is integrated into our module for performance comparison. This demonstrates that the proposed explicit regularization, combined with implicit regularization, is irreplaceable.
   \item [3.]\textbf{Sensitivity:} \emph{Does it require repeated adjustment of hyperparameters?} In Sec. 4.4, we discuss the sensitivity analysis of the hyperparameter $\delta$ and validate that FreLE is not sensitive to hyperparameters.
    \item [4.]\textbf{Efficiency:} \emph{Is it effective when reducing the number of parameters?} In Sec. 4.5, performance variation curves with different parameter quantities are presented, demonstrating that FreLE's method can be effectively utilized in stringent computational environments by reducing the number of parameters while maintaining strong performance.
\end{itemize}
\subsection{Set Up}

\subsubsection{\textbf{Baselines.}} In our experiments, the comparison baselines we adopted are primarily drawn from studies published between 2020 and 2024. These models can be categorized into three main groups based on their architectures: 1) Methods based on MLP: DLinear~\cite{DLinear}, RLinear~\cite{RLinear}, TiDE~\cite{Tide}, FreTS~\cite{FreTS}; 2) Methods based on the Transformer architecture: Autoformer~\cite{Auto}, FEDformer~\cite{Fedformer}, Fredformer~\cite{Fred}, iTransformer~\cite{itrans}, Stationary~\cite{Stationary}, TimesX~\cite{TimeX}; 3) Other well-known models: TimesNet~\cite{Timenet}.

\subsubsection{\textbf{Datasets.}} The datasets used for long-term forecasting include: ETT (h1, h2, m1, m2), Weather, Traffic, and Electricity~\cite{Data,LIB}. The information these datasets provide is summarized in Table~\ref{tab:database}.
\begin{table}[b]
    \caption{Benchmark dataset summary}
    \centering
    \resizebox{1\linewidth}{!}{
    \begin{tabular}{c|cccccccc}
    \toprule
        Datasets  &  Weather  &  Electricity & ETTh1 & ETTh2 & ETTm1 & ETTm2  & Traffic \\ 
            \midrule
\#Frequency&10min&Hourly&Hourly&Hourly&15min&15min&Hourly\\
        \#Channel & 21 & 321 & 7 & 7 & 7 & 7  & 862 \\
        \#D&21&321&7&7&7&7&862\\
        \#Timesteps  & 52969 & 26304 & 17420 & 17420 & 69680 & 69680   & 17544 \\ \bottomrule
    \end{tabular}}
    \label{tab:database}
\end{table}

\subsubsection{\textbf{Implementation.}} Regarding the reproduction of the baseline, it is based on the script of TimesNet\cite{Timenet} and FreDF\cite{FreDF}. Our experiments are conducted on GPU RTX 4090 and CPU with 14 cores, AMD EPYC 7453. The FreLE loss module is inserted into the DLinear model.
\subsection{Result}
Table~\ref{tablee} presents the prediction performance of different models across four selected datasets, with an input sequence length of 96 and prediction lengths of 96, 192, 336, and 720. In the complete set of seven datasets, FreLE achieved 21 first-place rankings and 17 second-place rankings.
\begin{table*}[htbp]
\scriptsize
  \caption{Multivariate forecasting results with prediction lengths $S \in\{96, 192, 336, 720\}$ for all datasets and fixed look-back length$T=96$. Experimental results for some datasets, with the \textbf{best} and \underline{second best} results are highlighted.}
  \vskip -0.0in
  \vspace{3pt}
  \renewcommand{\arraystretch}{0.85} 
  \centering
  \renewcommand{\multirowsetup}{\centering}
  \setlength{\tabcolsep}{1pt}
  \begin{tabular}{c|c|cc|cc|cc|cc|cc|cc|cc|cc|cc|cc|cc|cc}
    \toprule
    \multicolumn{2}{c}{\multirow{2}{*}{Models}} & 
    \multicolumn{2}{c}{\rotatebox{0}{\scalebox{0.8}{\textbf{\method}}}} &
    \multicolumn{2}{c}{\rotatebox{0}{\scalebox{0.8}{iTransformer}}} &
    \multicolumn{2}{c}{\rotatebox{0}{\scalebox{0.8}{RLinear}}} &
    \multicolumn{2}{c}{\rotatebox{0}{\scalebox{0.8}{Fredformer}}} &
    \multicolumn{2}{c}{\rotatebox{0}{\scalebox{0.8}{Crossformer}}}  &
    \multicolumn{2}{c}{\rotatebox{0}{\scalebox{0.8}{TiDE}}} &
    \multicolumn{2}{c}{\rotatebox{0}{\scalebox{0.8}{{TimesNet}}}} &
    \multicolumn{2}{c}{\rotatebox{0}{\scalebox{0.8}{DLinear}}}&
    \multicolumn{2}{c}{\rotatebox{0}{\scalebox{0.8}{FreTS}}} &
    \multicolumn{2}{c}{\rotatebox{0}{\scalebox{0.8}{FEDformer}}} &
    \multicolumn{2}{c}{\rotatebox{0}{\scalebox{0.8}{Stationary}}} &
    \multicolumn{2}{c}{\rotatebox{0}{\scalebox{0.8}{Autoformer}}} \\
    \multicolumn{2}{c}{} &
    \multicolumn{2}{c}{\scalebox{0.8}{\textbf{(Ours)}}} & 
   \multicolumn{2}{c}{\scalebox{0.8}{2023}} & 
    \multicolumn{2}{c}{\scalebox{0.8}{2023}} & 
    \multicolumn{2}{c}{\scalebox{0.8}{2024}} & 
    \multicolumn{2}{c}{\scalebox{0.8}{2023}}  & 
    \multicolumn{2}{c}{\scalebox{0.8}{2023}} & 
    \multicolumn{2}{c}{\scalebox{0.8}{2022}} & 
    \multicolumn{2}{c}{\scalebox{0.8}{2023}}& 
    \multicolumn{2}{c}{\scalebox{0.8}{2022}} &
    \multicolumn{2}{c}{\scalebox{0.8}{2022}} &
    \multicolumn{2}{c}{\scalebox{0.8}{2022}} &
    \multicolumn{2}{c}{\scalebox{0.8}{2021}} \\
    
    \cmidrule(lr){3-4} \cmidrule(lr){5-6}\cmidrule(lr){7-8} \cmidrule(lr){9-10}\cmidrule(lr){11-12}\cmidrule(lr){13-14} \cmidrule(lr){15-16} \cmidrule(lr){17-18} \cmidrule(lr){19-20} \cmidrule(lr){21-22} \cmidrule(lr){23-24}\cmidrule(lr){25-26}
    
    \multicolumn{2}{c}{Metric}  & \scalebox{0.78}{MSE} & \scalebox{0.78}{MAE} & \scalebox{0.78}{MSE} & \scalebox{0.78}{MAE}  & \scalebox{0.78}{MSE} & \scalebox{0.78}{MAE}  & \scalebox{0.78}{MSE} & \scalebox{0.78}{MAE} & \scalebox{0.78}{MSE} & \scalebox{0.78}{MAE}  & \scalebox{0.78}{MSE} & \scalebox{0.78}{MAE}  & \scalebox{0.78}{MSE} & \scalebox{0.78}{MAE} & \scalebox{0.78}{MSE} & \scalebox{0.78}{MAE} & \scalebox{0.78}{MSE} & \scalebox{0.78}{MAE} & \scalebox{0.78}{MSE} & \scalebox{0.78}{MAE} & \scalebox{0.78}{MSE} & \scalebox{0.78}{MAE} & \scalebox{0.78}{MSE} & \scalebox{0.78}{MAE} \\
    
    \toprule
    
    \multirow{5}{*}{\rotatebox{90}{\scalebox{0.95}{ETTm1}}}
    &  \scalebox{0.78}{96} & \boldres{\scalebox{0.78}{0.319}} & \boldres{\scalebox{0.78}{0.356}} & \scalebox{0.78}{0.334} & \scalebox{0.78}{0.368} & \scalebox{0.78}{0.355} & \scalebox{0.78}{0.376} & \secondres{\scalebox{0.78}{0.326}} & \secondres{\scalebox{0.78}{0.361}} & \scalebox{0.78}{0.404} & \scalebox{0.78}{0.426} & \scalebox{0.78}{0.364} & \scalebox{0.78}{0.387} &{\scalebox{0.78}{0.338}} &{\scalebox{0.78}{0.375}} &{\scalebox{0.78}{0.345}} &{\scalebox{0.78}{0.372}} & \scalebox{0.78}{0.339} & \scalebox{0.78}{0.374} &\scalebox{0.78}{0.379} &\scalebox{0.78}{0.419} &\scalebox{0.78}{0.386} &\scalebox{0.78}{0.398} &\scalebox{0.78}{0.505} &\scalebox{0.78}{0.475} \\ 
    
    & \scalebox{0.78}{192} & \secondres{\scalebox{0.78}{0.371}} & \boldres{\scalebox{0.78}{0.379}} & \scalebox{0.78}{0.377} & \scalebox{0.78}{0.391} & \scalebox{0.78}{0.391} & \scalebox{0.78}{0.392} & \boldres{\scalebox{0.78}{0.363}} & \secondres{\scalebox{0.78}{0.380}} & \scalebox{0.78}{0.450} & \scalebox{0.78}{0.451} &\scalebox{0.78}{0.398} & \scalebox{0.78}{0.404} &\scalebox{0.78}{0.374} & \scalebox{0.78}{0.387} &{\scalebox{0.78}{0.380}} &{\scalebox{0.78}{0.389}} & \scalebox{0.78}{0.382} & \scalebox{0.78}{0.397}  &\scalebox{0.78}{0.426} &\scalebox{0.78}{0.441} &\scalebox{0.78}{0.459} &\scalebox{0.78}{0.444} &\scalebox{0.78}{0.553} &\scalebox{0.78}{0.496} \\ 
    
    & \scalebox{0.78}{336} &\boldres{ \scalebox{0.78}{0.393}} & \boldres{\scalebox{0.78}{0.401}} & \scalebox{0.78}{0.426} & \scalebox{0.78}{0.420} & \scalebox{0.78}{0.424} & \scalebox{0.78}{0.415} & \secondres{\scalebox{0.78}{0.395}} & \secondres{\scalebox{0.78}{0.403}} & \scalebox{0.78}{0.532}  &\scalebox{0.78}{0.515} & \scalebox{0.78}{0.428} & \scalebox{0.78}{0.425} & \scalebox{0.78}{0.410} & \scalebox{0.78}{0.411}  &{\scalebox{0.78}{0.413}} &{\scalebox{0.78}{0.413}} & \scalebox{0.78}{0.421} & \scalebox{0.78}{0.426}  &\scalebox{0.78}{0.445} &\scalebox{0.78}{0.459} &\scalebox{0.78}{0.495} &\scalebox{0.78}{0.464} &\scalebox{0.78}{0.621} &\scalebox{0.78}{0.537} \\ 
    
    & \scalebox{0.78}{720} & \secondres{\scalebox{0.78}{0.463}} & \secondres{\scalebox{0.78}{0.443}} & \scalebox{0.78}{0.491} & \scalebox{0.78}{0.459} & \scalebox{0.78}{0.487} & \scalebox{0.78}{0.450} & \boldres{\scalebox{0.78}{0.453}} & \boldres{\scalebox{0.78}{0.438}} & \scalebox{0.78}{0.666} & \scalebox{0.78}{0.589} & \scalebox{0.78}{0.487} & \scalebox{0.78}{0.461} &{\scalebox{0.78}{0.478}} & \scalebox{0.78}{0.450} & \scalebox{0.78}{0.474} &{\scalebox{0.78}{0.453}} & \scalebox{0.78}{0.485} & \scalebox{0.78}{0.462}  &\scalebox{0.78}{0.543} &\scalebox{0.78}{0.490} &\scalebox{0.78}{0.585} &\scalebox{0.78}{0.516} &\scalebox{0.78}{0.671} &\scalebox{0.78}{0.561} \\ 
    
    \cmidrule(lr){2-26}
    
    & \scalebox{0.78}{Avg} & \boldres{\scalebox{0.78}{0.386}} & \boldres{\scalebox{0.78}{0.394}} & \scalebox{0.78}{0.407} & \scalebox{0.78}{0.410} & \scalebox{0.78}{0.414} & \scalebox{0.78}{0.407} & \secondres{\scalebox{0.78}{0.387}} & \secondres{\scalebox{0.78}{0.400}} & \scalebox{0.78}{0.513} & \scalebox{0.78}{0.496} & \scalebox{0.78}{0.419} & \scalebox{0.78}{0.419} & \scalebox{0.78}{0.400} & \scalebox{0.78}{0.406}  &{\scalebox{0.78}{0.403}} &{\scalebox{0.78}{0.407}} & \scalebox{0.78}{0.407} & \scalebox{0.78}{0.415}  &\scalebox{0.78}{0.448} &\scalebox{0.78}{0.452} &\scalebox{0.78}{0.481} &\scalebox{0.78}{0.456} &\scalebox{0.78}{0.588} &\scalebox{0.78}{0.517} \\ 
    
    \midrule
    
    \multirow{5}{*}{\rotatebox{90}{\scalebox{0.95}{ETTm2}}}
    &  \scalebox{0.78}{96} & \boldres{\scalebox{0.78}{0.174}} & \boldres{\scalebox{0.78}{0.256}} & \scalebox{0.78}{0.180} & \scalebox{0.78}{0.264} & \scalebox{0.78}{0.182} & \scalebox{0.78}{0.265} & \secondres{\scalebox{0.78}{0.177}} & \secondres{\scalebox{0.78}{0.259}} & \scalebox{0.78}{0.287} & \scalebox{0.78}{0.366} & \scalebox{0.78}{0.207} & \scalebox{0.78}{0.305} &{\scalebox{0.78}{0.187}} &\scalebox{0.78}{0.267} &\scalebox{0.78}{0.193} &\scalebox{0.78}{0.292} & \scalebox{0.78}{0.190} & \scalebox{0.78}{0.282} &\scalebox{0.78}{0.203} &\scalebox{0.78}{0.287} &{\scalebox{0.78}{0.192}} &\scalebox{0.78}{0.274} &\scalebox{0.78}{0.255} &\scalebox{0.78}{0.339} \\ 
    
    & \scalebox{0.78}{192} & \boldres{\scalebox{0.78}{0.239}} & \boldres{\scalebox{0.78}{0.295}} & \scalebox{0.78}{0.250} & \scalebox{0.78}{0.309} & \scalebox{0.78}{0.246} & \scalebox{0.78}{0.304} & \secondres{\scalebox{0.78}{0.243}} & \secondres{\scalebox{0.78}{0.301}} & \scalebox{0.78}{0.414} & \scalebox{0.78}{0.492} & \scalebox{0.78}{0.290} & \scalebox{0.78}{0.364} &{\scalebox{0.78}{0.249}} &{\scalebox{0.78}{0.309}} &\scalebox{0.78}{0.284} &\scalebox{0.78}{0.362} & \scalebox{0.78}{0.260} & \scalebox{0.78}{0.329} &\scalebox{0.78}{0.269} &\scalebox{0.78}{0.328} &\scalebox{0.78}{0.280} &\scalebox{0.78}{0.339} &\scalebox{0.78}{0.281} &\scalebox{0.78}{0.340} \\ 
    
    & \scalebox{0.78}{336} & \boldres{\scalebox{0.78}{0.300}} & \boldres{\scalebox{0.78}{0.334}} & {\scalebox{0.78}{0.311}} & {\scalebox{0.78}{0.348}} & \scalebox{0.78}{0.307} & {\scalebox{0.78}{0.342}} & \secondres{\scalebox{0.78}{0.302}} & \secondres{\scalebox{0.78}{0.340}}  & \scalebox{0.78}{0.597} & \scalebox{0.78}{0.542}  & \scalebox{0.78}{0.377} & \scalebox{0.78}{0.422} &{\scalebox{0.78}{0.321}} &{\scalebox{0.78}{0.351}} &\scalebox{0.78}{0.369} &\scalebox{0.78}{0.427} & \scalebox{0.78}{0.373} & \scalebox{0.78}{0.405} &\scalebox{0.78}{0.325} &\scalebox{0.78}{0.366} &\scalebox{0.78}{0.334} &\scalebox{0.78}{0.361} &\scalebox{0.78}{0.339} &\scalebox{0.78}{0.372} \\ 
    
    & \scalebox{0.78}{720} & \boldres{\scalebox{0.78}{0.399}} & \boldres{\scalebox{0.78}{0.392}} & \scalebox{0.78}{0.412} & \scalebox{0.78}{0.407} & \scalebox{0.78}{0.407} & {\scalebox{0.78}{0.398}} & \secondres{\scalebox{0.78}{0.397}} & \secondres{\scalebox{0.78}{0.396}} & \scalebox{0.78}{1.730} & \scalebox{0.78}{1.042} & \scalebox{0.78}{0.558} & \scalebox{0.78}{0.524} &{\scalebox{0.78}{0.408}} &{\scalebox{0.78}{0.403}} &\scalebox{0.78}{0.554} &\scalebox{0.78}{0.522} & \scalebox{0.78}{0.517} & \scalebox{0.78}{0.499} &\scalebox{0.78}{0.421} &\scalebox{0.78}{0.415} &\scalebox{0.78}{0.417} &\scalebox{0.78}{0.413} &\scalebox{0.78}{0.433} &\scalebox{0.78}{0.432} \\ 
    
    \cmidrule(lr){2-26}
    
    & \scalebox{0.78}{Avg} & \boldres{\scalebox{0.78}{0.278}} & \boldres{\scalebox{0.78}{0.319}} & {\scalebox{0.78}{0.288}} & {\scalebox{0.78}{0.332}} & \scalebox{0.78}{0.286} & \scalebox{0.78}{0.327} & \secondres{\scalebox{0.78}{0.279}} & \secondres{\scalebox{0.78}{0.324}} & \scalebox{0.78}{0.757} & \scalebox{0.78}{0.610} & \scalebox{0.78}{0.358} & \scalebox{0.78}{0.404} &{\scalebox{0.78}{0.291}} &{\scalebox{0.78}{0.333}} &\scalebox{0.78}{0.350} &\scalebox{0.78}{0.401} & \scalebox{0.78}{0.335} & \scalebox{0.78}{0.379} &\scalebox{0.78}{0.305} &\scalebox{0.78}{0.349} &\scalebox{0.78}{0.306} &\scalebox{0.78}{0.347} &\scalebox{0.78}{0.327} &\scalebox{0.78}{0.371} \\ 
    
    \midrule
    
    \multirow{5}{*}{\rotatebox{90}{
    \scalebox{0.95}{ETTh1}}}
    &  \scalebox{0.78}{96} & \boldres{\scalebox{0.78}{0.371}} & \boldres{\scalebox{0.78}{0.392}} & {\scalebox{0.78}{0.386}} & {\scalebox{0.78}{0.405}} & \scalebox{0.78}{0.386} & \secondres{\scalebox{0.78}{0.395}} & \secondres{\scalebox{0.78}{0.373}} & \boldres{\scalebox{0.78}{0.392}} & \scalebox{0.78}{0.423} & \scalebox{0.78}{0.448} & \scalebox{0.78}{0.479}& \scalebox{0.78}{0.464}  & \scalebox{0.78}{0.384} &{\scalebox{0.78}{0.412}} & \scalebox{0.78}{0.386} & \scalebox{0.78}{0.400} & \scalebox{0.78}{0.399} & \scalebox{0.78}{0.433} & \scalebox{0.78}{0.376} &\scalebox{0.78}{0.419} &\scalebox{0.78}{0.513} &\scalebox{0.78}{0.491} &\scalebox{0.78}{0.449} &\scalebox{0.78}{0.459}  \\ 
    
    & \scalebox{0.78}{192} & \boldres{\scalebox{0.78}{0.425}} & \secondres{\scalebox{0.78}{0.423}} & \scalebox{0.78}{0.441} & \scalebox{0.78}{0.436} & {\scalebox{0.78}{0.437}} & {\scalebox{0.78}{0.424}} & \secondres{\scalebox{0.78}{0.433}} & \boldres{\scalebox{0.78}{0.420}} & \scalebox{0.78}{0.471} & \scalebox{0.78}{0.474}  & \scalebox{0.78}{0.525} & \scalebox{0.78}{0.492} &\scalebox{0.78}{0.436} & \scalebox{0.78}{0.429}  &{\scalebox{0.78}{0.437}} &{\scalebox{0.78}{0.432}} & \scalebox{0.78}{0.453} & \scalebox{0.78}{0.443} &{\scalebox{0.78}{0.437}} &\scalebox{0.78}{0.448} &\scalebox{0.78}{0.534} &\scalebox{0.78}{0.504} &\scalebox{0.78}{0.500} &\scalebox{0.78}{0.482} \\ 
    
    & \scalebox{0.78}{336} & \boldres{\scalebox{0.78}{0.467}} & \secondres{\scalebox{0.78}{0.445}} & \scalebox{0.78}{0.487} & \scalebox{0.78}{0.458} & \scalebox{0.78}{0.479} & \scalebox{0.78}{0.446} & \secondres{\scalebox{0.78}{0.470}} & \boldres{\scalebox{0.78}{0.437}} & \scalebox{0.78}{0.570} & \scalebox{0.78}{0.546} & \scalebox{0.78}{0.565} & \scalebox{0.78}{0.515} &\scalebox{0.78}{0.491} &\scalebox{0.78}{0.469} &{\scalebox{0.78}{0.481}} & \scalebox{0.78}{0.459} & \scalebox{0.78}{0.503} & \scalebox{0.78}{0.475} &{\scalebox{0.78}{0.479}} &\scalebox{0.78}{0.465} &\scalebox{0.78}{0.588} &\scalebox{0.78}{0.535} &\scalebox{0.78}{0.521} &\scalebox{0.78}{0.496} \\ 
    
    & \scalebox{0.78}{720} & \secondres{\scalebox{0.78}{0.476}} & \secondres{\scalebox{0.78}{0.473}} & {\scalebox{0.78}{0.503}} & {\scalebox{0.78}{0.491}} & \boldres{\scalebox{0.78}{0.456}} & 
    \boldres{\scalebox{0.78}{0.470}} & \secondres{\scalebox{0.78}{0.467}} & \scalebox{0.78}{0.488} & \scalebox{0.78}{0.653} & \scalebox{0.78}{0.621} & \scalebox{0.78}{0.594} & \scalebox{0.78}{0.558} &\scalebox{0.78}{0.521} &{\scalebox{0.78}{0.500}} &\scalebox{0.78}{0.519} &\scalebox{0.78}{0.516} & \scalebox{0.78}{0.596} & \scalebox{0.78}{0.565} &{\scalebox{0.78}{0.506}} &{\scalebox{0.78}{0.507}} &\scalebox{0.78}{0.643} &\scalebox{0.78}{0.616} &{\scalebox{0.78}{0.514}} &\scalebox{0.78}{0.512}  \\ 
    
    \cmidrule(lr){2-26}
    
    & \scalebox{0.78}{Avg} & \boldres{\scalebox{0.78}{0.434}} & \secondres{\scalebox{0.78}{0.433}} & {\scalebox{0.78}{0.454}} & \scalebox{0.78}{0.447} & \secondres{\scalebox{0.78}{0.435}} & \secondres{\scalebox{0.78}{0.433}} & \secondres{\scalebox{0.78}{0.435}} & \boldres{\scalebox{0.78}{0.426}} & \scalebox{0.78}{0.529} & \scalebox{0.78}{0.522} & \scalebox{0.78}{0.541} & \scalebox{0.78}{0.507} &\scalebox{0.78}{0.458} &{\scalebox{0.78}{0.450}} &{\scalebox{0.78}{0.456}} &{\scalebox{0.78}{0.452}} & \scalebox{0.78}{0.488} & \scalebox{0.78}{0.474} &{\scalebox{0.78}{0.450}} &\scalebox{0.78}{0.460} &\scalebox{0.78}{0.570} &\scalebox{0.78}{0.537} &\scalebox{0.78}{0.496} &\scalebox{0.78}{0.487}  \\ 
    
    \midrule

    \multirow{5}{*}{\rotatebox{90}{\scalebox{0.95}{ETTh2}}}
    &  \scalebox{0.78}{96} & \secondres{\scalebox{0.78}{0.284}} & \secondres{\scalebox{0.78}{0.336}} & \scalebox{0.78}{0.297} & {\scalebox{0.78}{0.349}} & \boldres{\scalebox{0.78}{0.288}} & \boldres{\scalebox{0.78}{0.338}} & {\scalebox{0.78}{0.293}} & \scalebox{0.78}{0.342} & \scalebox{0.78}{0.745} & \scalebox{0.78}{0.584} &\scalebox{0.78}{0.400} & \scalebox{0.78}{0.440}  & {\scalebox{0.78}{0.340}} & {\scalebox{0.78}{0.374}} &{\scalebox{0.78}{0.333}} &{\scalebox{0.78}{0.387}} & \scalebox{0.78}{0.350} & \scalebox{0.78}{0.403}  &\scalebox{0.78}{0.358} &\scalebox{0.78}{0.397} &\scalebox{0.78}{0.476} &\scalebox{0.78}{0.458} &\scalebox{0.78}{0.346} &\scalebox{0.78}{0.388} \\ 
    
    & \scalebox{0.78}{192} & \boldres{\scalebox{0.78}{0.370}} & \boldres{\scalebox{0.78}{0.388}} & \scalebox{0.78}{0.380} & \scalebox{0.78}{0.400}&{\scalebox{0.78}{0.374}} & {\scalebox{0.78}{0.390}} & \secondres{\scalebox{0.78}{0.371}} & \secondres{\scalebox{0.78}{0.389}} & \scalebox{0.78}{0.877} & \scalebox{0.78}{0.656} & \scalebox{0.78}{0.528} & \scalebox{0.78}{0.509} & {\scalebox{0.78}{0.402}} & {\scalebox{0.78}{0.414}} &\scalebox{0.78}{0.477} &\scalebox{0.78}{0.476} & \scalebox{0.78}{0.472} & \scalebox{0.78}{0.475} &{\scalebox{0.78}{0.429}} &{\scalebox{0.78}{0.439}} &\scalebox{0.78}{0.512} &\scalebox{0.78}{0.493} &\scalebox{0.78}{0.456} &\scalebox{0.78}{0.452} \\ 
    
    & \scalebox{0.78}{336} & \secondres{\scalebox{0.78}{0.413}} & \boldres{\scalebox{0.78}{0.403}} & {\scalebox{0.78}{0.428}} & \scalebox{0.78}{0.432} & {\scalebox{0.78}{0.415}} & {\scalebox{0.78}{0.426}} & \boldres{\scalebox{0.78}{0.382}} & \secondres{\scalebox{0.78}{0.409}}& \scalebox{0.78}{1.043} & \scalebox{0.78}{0.731} & \scalebox{0.78}{0.643} & \scalebox{0.78}{0.571}  & {\scalebox{0.78}{0.452}} & {\scalebox{0.78}{0.452}} &\scalebox{0.78}{0.594} &\scalebox{0.78}{0.541} & \scalebox{0.78}{0.564} &\scalebox{0.78}{0.528} &\scalebox{0.78}{0.496} &\scalebox{0.78}{0.487} &\scalebox{0.78}{0.552} &\scalebox{0.78}{0.551} &{\scalebox{0.78}{0.482}} &\scalebox{0.78}{0.486}\\ 
    
    & \scalebox{0.78}{720} & \boldres{\scalebox{0.78}{0.412}} & \boldres{\scalebox{0.78}{0.429}} & \scalebox{0.78}{0.427} & \scalebox{0.78}{0.445} & \secondres{\scalebox{0.78}{0.415}} & \secondres{\scalebox{0.78}{0.434}} & {\scalebox{0.78}{0.431}} & {\scalebox{0.78}{0.446}} & \scalebox{0.78}{1.104} & \scalebox{0.78}{0.763} & \scalebox{0.78}{0.874} & \scalebox{0.78}{0.679} & {\scalebox{0.78}{0.462}} & {\scalebox{0.78}{0.468}} &\scalebox{0.78}{0.831} &\scalebox{0.78}{0.657} & \scalebox{0.78}{0.815} & \scalebox{0.78}{0.654} &{\scalebox{0.78}{0.463}} &{\scalebox{0.78}{0.474}} &\scalebox{0.78}{0.562} &\scalebox{0.78}{0.560} &\scalebox{0.78}{0.515} &\scalebox{0.78}{0.511} \\ 
    
    \cmidrule(lr){2-26}
    
    & \scalebox{0.78}{Avg} & \secondres{\scalebox{0.78}{0.370}} & \boldres{\scalebox{0.78}{0.389}} & \scalebox{0.78}{0.383} & \scalebox{0.78}{0.407} & {\scalebox{0.78}{0.374}} & {\scalebox{0.78}{0.398}} & \boldres{\scalebox{0.78}{0.365}} & \secondres{\scalebox{0.78}{0.393}} & \scalebox{0.78}{0.942} & \scalebox{0.78}{0.684} & \scalebox{0.78}{0.611} & \scalebox{0.78}{0.550}  &{\scalebox{0.78}{0.414}} &{\scalebox{0.78}{0.427}} &\scalebox{0.78}{0.559} &\scalebox{0.78}{0.515} & \scalebox{0.78}{0.550} & \scalebox{0.78}{0.515} &\scalebox{0.78}{{0.437}} &\scalebox{0.78}{{0.449}} &\scalebox{0.78}{0.526} &\scalebox{0.78}{0.516} &\scalebox{0.78}{0.450} &\scalebox{0.78}{0.459} \\ 
    
    \midrule
    
    \multirow{5}{*}{\rotatebox{90}{\scalebox{0.95}{ECL}}} 
    &  \scalebox{0.78}{96} & \boldres{\scalebox{0.78}{0.144}} & {\scalebox{0.78}{0.249}} & {\scalebox{0.78}{0.148}} & \boldres{\scalebox{0.78}{0.240}} & \scalebox{0.78}{0.201} & \scalebox{0.78}{0.281} & \secondres{\scalebox{0.78}{0.147}} & \secondres{\scalebox{0.78}{0.241}} & \scalebox{0.78}{0.219} & \scalebox{0.78}{0.314} & \scalebox{0.78}{0.237} & \scalebox{0.78}{0.329} &\scalebox{0.78}{0.168} &\scalebox{0.78}{0.272} &\scalebox{0.78}{0.197} &\scalebox{0.78}{0.282} & \scalebox{0.78}{0.189} & \scalebox{0.78}{0.277} &\scalebox{0.78}{0.193} &\scalebox{0.78}{0.308} &{\scalebox{0.78}{0.169}} &{\scalebox{0.78}{0.273}} &\scalebox{0.78}{0.201} &\scalebox{0.78}{0.317}  \\ 
    
    & \scalebox{0.78}{192} & \boldres{\scalebox{0.78}{0.162}} & \secondres{\scalebox{0.78}{0.255}} & \boldres{\scalebox{0.78}{0.162}} & \boldres{\scalebox{0.78}{0.253}} & \scalebox{0.78}{0.201} & \scalebox{0.78}{0.283} & \secondres{\scalebox{0.78}{0.165}} & \scalebox{0.78}{0.258} & \scalebox{0.78}{0.231} & \scalebox{0.78}{0.322} & \scalebox{0.78}{0.236} & \scalebox{0.78}{0.330} &{\scalebox{0.78}{0.184}} &\scalebox{0.78}{0.289} &\scalebox{0.78}{0.196} &{\scalebox{0.78}{0.285}} & \scalebox{0.78}{0.193} & \scalebox{0.78}{0.282} &\scalebox{0.78}{0.201} &\scalebox{0.78}{0.315} &\scalebox{0.78}{0.182} &\scalebox{0.78}{0.286} &\scalebox{0.78}{0.222} &\scalebox{0.78}{0.334} \\ 
    
    & \scalebox{0.78}{336} & {\scalebox{0.78}{0.179}} & {\scalebox{0.78}{0.285}} & \secondres{\scalebox{0.78}{0.178}} & \boldres{\scalebox{0.78}{0.269}} & \boldres{\scalebox{0.78}{0.177}} & \secondres{\scalebox{0.78}{0.273}}& \scalebox{0.78}{0.181} & \scalebox{0.78}{0.305} & \scalebox{0.78}{0.246} & \scalebox{0.78}{0.337} & \scalebox{0.78}{0.249} & \scalebox{0.78}{0.344} &\scalebox{0.78}{0.198} &{\scalebox{0.78}{0.300}} &\scalebox{0.78}{0.209} &{\scalebox{0.78}{0.301}} & \scalebox{0.78}{0.207} & \scalebox{0.78}{0.296} &\scalebox{0.78}{0.214} &\scalebox{0.78}{0.329} &{\scalebox{0.78}{0.200}} &\scalebox{0.78}{0.304} &\scalebox{0.78}{0.231} &\scalebox{0.78}{0.338}  \\ 
    
    & \scalebox{0.78}{720} & \boldres{\scalebox{0.78}{0.213}} & \boldres{\scalebox{0.78}{0.296}}  & \scalebox{0.78}{0.225} & {\scalebox{0.78}{0.317}} & \scalebox{0.78}{0.257} & \scalebox{0.78}{0.331} & \boldres{\scalebox{0.78}{0.213}} & \secondres{\scalebox{0.78}{0.304}} & \scalebox{0.78}{0.280} & \scalebox{0.78}{0.363} & \scalebox{0.78}{0.284} & \scalebox{0.78}{0.373} &\secondres{\scalebox{0.78}{0.220}} &\scalebox{0.78}{0.320}&\scalebox{0.78}{0.245} &\scalebox{0.78}{0.333} & \scalebox{0.78}{0.245} & \scalebox{0.78}{0.332} &\scalebox{0.78}{0.246} &\scalebox{0.78}{0.355} &{\scalebox{0.78}{0.222}} &{\scalebox{0.78}{0.321}} &\scalebox{0.78}{0.254} &\scalebox{0.78}{0.361} \\ 
    
    \cmidrule(lr){2-26}
    
    & \scalebox{0.78}{Avg} & \boldres{\scalebox{0.78}{0.175}} & {\scalebox{0.78}{0.271}} & {\scalebox{0.78}{0.178}} & \secondres{\scalebox{0.78}{0.270}} & \scalebox{0.78}{0.219} & \scalebox{0.78}{0.298} & \secondres{\scalebox{0.78}{0.177}} & \boldres{\scalebox{0.78}{0.269}} & \scalebox{0.78}{0.244} & \scalebox{0.78}{0.334} & \scalebox{0.78}{0.251} & \scalebox{0.78}{0.344} &\scalebox{0.78}{0.192} &\scalebox{0.78}{0.295} &\scalebox{0.78}{0.212} &\scalebox{0.78}{0.300} & \scalebox{0.78}{0.209} & \scalebox{0.78}{0.297} &\scalebox{0.78}{0.214} &\scalebox{0.78}{0.327} &{\scalebox{0.78}{0.193}} &{\scalebox{0.78}{0.296}} &\scalebox{0.78}{0.227} &\scalebox{0.78}{0.338} \\ 
    
    \midrule
    
    \multirow{5}{*}{\rotatebox{90}{\scalebox{0.95}{Traffic}}} 
    & \scalebox{0.78}{96} & {\scalebox{0.78}{0.412}} & {\scalebox{0.78}{0.294}} & \boldres{\scalebox{0.78}{0.395}} & \boldres{\scalebox{0.78}{0.268}} & \scalebox{0.78}{0.649} & \scalebox{0.78}{0.389} & \secondres{\scalebox{0.78}{0.406}} & \secondres{\scalebox{0.78}{0.277}} & \scalebox{0.78}{0.522} & \scalebox{0.78}{0.290} & \scalebox{0.78}{0.805} & \scalebox{0.78}{0.493} &{\scalebox{0.78}{0.593}} &{\scalebox{0.78}{0.321}} &\scalebox{0.78}{0.650} &\scalebox{0.78}{0.396} & \scalebox{0.78}{0.528} & \scalebox{0.78}{0.341} &{\scalebox{0.78}{0.587}} &\scalebox{0.78}{0.366} &\scalebox{0.78}{0.612} &{\scalebox{0.78}{0.338}} &\scalebox{0.78}{0.613} &\scalebox{0.78}{0.388} \\ 
    
    & \scalebox{0.78}{192} & {\scalebox{0.78}{0.437}} & \secondres{\scalebox{0.78}{0.286}} & \boldres{\scalebox{0.78}{0.417}} & \boldres{\scalebox{0.78}{0.276}} & \scalebox{0.78}{0.601} & \scalebox{0.78}{0.366} & \secondres{\scalebox{0.78}{0.426}} & \scalebox{0.78}{0.290} & \scalebox{0.78}{0.530} & \scalebox{0.78}{0.293} & \scalebox{0.78}{0.756} & \scalebox{0.78}{0.474} &\scalebox{0.78}{0.617} &{\scalebox{0.78}{0.336}} &{\scalebox{0.78}{0.598}} &\scalebox{0.78}{0.370} & \scalebox{0.78}{0.193} & \scalebox{0.78}{0.282} &\scalebox{0.78}{0.604} &\scalebox{0.78}{0.373} &\scalebox{0.78}{0.613} &{\scalebox{0.78}{0.340}} &\scalebox{0.78}{0.616} &\scalebox{0.78}{0.382}  \\ 
    
    & \scalebox{0.78}{336} & {\scalebox{0.78}{0.438}} &\secondres{\scalebox{0.78}{0.282}} & \secondres{\scalebox{0.78}{0.433}} & {\scalebox{0.78}{0.283}} & \scalebox{0.78}{0.609} & \scalebox{0.78}{0.369} & \boldres{\scalebox{0.78}{0.432}} &  \boldres{\scalebox{0.78}{0.281}} & \scalebox{0.78}{0.558} & \scalebox{0.78}{0.305}  & \scalebox{0.78}{0.762} & \scalebox{0.78}{0.477} &\scalebox{0.78}{0.629} &{\scalebox{0.78}{0.336}}  &{\scalebox{0.78}{0.605}} &\scalebox{0.78}{0.373} & \scalebox{0.78}{0.551} & \scalebox{0.78}{0.345}&\scalebox{0.78}{0.621} &\scalebox{0.78}{0.383} &\scalebox{0.78}{0.618} &{\scalebox{0.78}{0.328}} &\scalebox{0.78}{0.622} &\scalebox{0.78}{0.337} \\ 
    
    & \scalebox{0.78}{720} & \boldres{\scalebox{0.78}{0.459}} & \boldres{\scalebox{0.78}{0.299}} & {\scalebox{0.78}{0.467}} & {\scalebox{0.78}{0.302}} & \scalebox{0.78}{0.647} & \scalebox{0.78}{0.387} & \secondres{\scalebox{0.78}{0.463}} & \secondres{\scalebox{0.78}{0.300}} & \scalebox{0.78}{0.589} & \scalebox{0.78}{0.328}  & \scalebox{0.78}{0.719} & \scalebox{0.78}{0.449} &\scalebox{0.78}{0.640} &{\scalebox{0.78}{0.350}} &\scalebox{0.78}{0.645} &\scalebox{0.78}{0.394} & \scalebox{0.78}{0.598} & \scalebox{0.78}{0.367} &{\scalebox{0.78}{0.626}} &\scalebox{0.78}{0.382} &\scalebox{0.78}{0.653} &{\scalebox{0.78}{0.355}} &\scalebox{0.78}{0.660} &\scalebox{0.78}{0.408} \\ 
    
    \cmidrule(lr){2-26}
    
    & \scalebox{0.78}{Avg} & {\scalebox{0.78}{0.436}} & {\scalebox{0.78}{0.290}} & \boldres{\scalebox{0.78}{0.428}} & \boldres{\scalebox{0.78}{0.282}} & \scalebox{0.78}{0.626} & \scalebox{0.78}{0.378} & \secondres{\scalebox{0.78}{0.431}} & \secondres{\scalebox{0.78}{0.287}} & \scalebox{0.78}{0.550} & \scalebox{0.78}{0.304} & \scalebox{0.78}{0.760} & \scalebox{0.78}{0.473} &{\scalebox{0.78}{0.620}} &{\scalebox{0.78}{0.336}} &\scalebox{0.78}{0.625} &\scalebox{0.78}{0.383} & \scalebox{0.78}{0.552} & \scalebox{0.78}{0.348} &{\scalebox{0.78}{0.610}} &\scalebox{0.78}{0.376} &\scalebox{0.78}{0.624} &{\scalebox{0.78}{0.340}} &\scalebox{0.78}{0.628} &\scalebox{0.78}{0.379} \\ 
    
    \midrule
    
    \multirow{5}{*}{\rotatebox{90}{\scalebox{0.95}{Weather}}} 
    &  \scalebox{0.78}{96} & {\scalebox{0.78}{0.171}} & \secondres{\scalebox{0.78}{0.212}} & \scalebox{0.78}{0.174} & {\scalebox{0.78}{0.214}} & \scalebox{0.78}{0.192} & \scalebox{0.78}{0.232} & \secondres{\scalebox{0.78}{0.163}} & \boldres{\scalebox{0.78}{0.207}} & \boldres{\scalebox{0.78}{0.158}} & \scalebox{0.78}{0.230}  & \scalebox{0.78}{0.202} & \scalebox{0.78}{0.261} &\scalebox{0.78}{0.172} &{\scalebox{0.78}{0.220}} & \scalebox{0.78}{0.196} &\scalebox{0.78}{0.255} & \scalebox{0.78}{0.184} & \scalebox{0.78}{0.239} & \scalebox{0.78}{0.217} &\scalebox{0.78}{0.296} & {\scalebox{0.78}{0.173}} &{\scalebox{0.78}{0.223}} & \scalebox{0.78}{0.266} &\scalebox{0.78}{0.336} \\ 
    
    & \scalebox{0.78}{192} & {\scalebox{0.78}{0.219}} & \boldres{\scalebox{0.78}{0.245}} & \scalebox{0.78}{0.221} & {\scalebox{0.78}{0.254}} & \scalebox{0.78}{0.240} & \scalebox{0.78}{0.271} & \secondres{\scalebox{0.78}{0.211}} & \secondres{\scalebox{0.78}{0.251}} & \boldres{\scalebox{0.78}{0.206}} & \scalebox{0.78}{0.277} & \scalebox{0.78}{0.242} & \scalebox{0.78}{0.298} &\scalebox{0.78}{0.219} &\scalebox{0.78}{0.223} & \scalebox{0.78}{0.275} &\scalebox{0.78}{0.296} & \scalebox{0.78}{0.261} & \scalebox{0.78}{0.340} & \scalebox{0.78}{0.276} &\scalebox{0.78}{0.336} & \scalebox{0.78}{0.245} &\scalebox{0.78}{0.285} & \scalebox{0.78}{0.307} &\scalebox{0.78}{0.367} \\ 
    
    & \scalebox{0.78}{336} & \boldres{\scalebox{0.78}{0.258}} & {\scalebox{0.78}{0.304}} & \scalebox{0.78}{0.278} & \secondres{\scalebox{0.78}{0.296}} & \scalebox{0.78}{0.292} & \scalebox{0.78}{0.307} & \secondres{\scalebox{0.78}{0.267}} & \boldres{\scalebox{0.78}{0.292}} & {\scalebox{0.78}{0.272}} & \scalebox{0.78}{0.335} & \scalebox{0.78}{0.287} & \scalebox{0.78}{0.335} &{\scalebox{0.78}{0.280}} &{\scalebox{0.78}{0.306}} & \scalebox{0.78}{0.283} &\scalebox{0.78}{0.335} & \scalebox{0.78}{0.272} & \scalebox{0.78}{0.316} & \scalebox{0.78}{0.339} &\scalebox{0.78}{0.380} & \scalebox{0.78}{0.321} &\scalebox{0.78}{0.338} & \scalebox{0.78}{0.359} &\scalebox{0.78}{0.395}\\ 
    
    & \scalebox{0.78}{720} & \boldres{\scalebox{0.78}{0.341}} & \secondres{\scalebox{0.78}{0.348}} & \scalebox{0.78}{0.358} & {\scalebox{0.78}{0.349}} & \scalebox{0.78}{0.364} & \scalebox{0.78}{0.353} & \secondres{\scalebox{0.78}{0.343}} & \boldres{\scalebox{0.78}{0.341}} & \scalebox{0.78}{0.398} & \scalebox{0.78}{0.418} & {\scalebox{0.78}{0.351}} & \scalebox{0.78}{0.386} &\scalebox{0.78}{0.365} &{\scalebox{0.78}{0.359}} & {\scalebox{0.78}{0.345}} &{\scalebox{0.78}{0.381}} & \scalebox{0.78}{0.340} & \scalebox{0.78}{0.363} & \scalebox{0.78}{0.403} &\scalebox{0.78}{0.428} & \scalebox{0.78}{0.414} &\scalebox{0.78}{0.410} & \scalebox{0.78}{0.419} &\scalebox{0.78}{0.428} \\ 
    
    \cmidrule(lr){2-26}
    
    & \scalebox{0.78}{Avg} & \secondres{\scalebox{0.78}{0.247}} & \secondres{\scalebox{0.78}{0.277}} & {\scalebox{0.78}{0.258}} & {\scalebox{0.78}{0.279}} & \scalebox{0.78}{0.272} & \scalebox{0.78}{0.291} & \boldres{\scalebox{0.78}{0.246}} & \boldres{\scalebox{0.78}{0.272}} & \scalebox{0.78}{0.259} & \scalebox{0.78}{0.315} & \scalebox{0.78}{0.271} & \scalebox{0.78}{0.320} &{\scalebox{0.78}{0.259}} &{\scalebox{0.78}{0.287}} &\scalebox{0.78}{0.265} &\scalebox{0.78}{0.317} & \scalebox{0.78}{0.255} & \scalebox{0.78}{0.299} &\scalebox{0.78}{0.309} &\scalebox{0.78}{0.360} &\scalebox{0.78}{0.288} &\scalebox{0.78}{0.314} &\scalebox{0.78}{0.338} &\scalebox{0.78}{0.382} \\ 

    \midrule
    
     \multicolumn{2}{c|}{\scalebox{0.78}{{$1^{\text{st}}$ Count}}} & \scalebox{0.78}{\boldres{21}} & \scalebox{0.78}{\boldres{17}} & \scalebox{0.78}{4} & \scalebox{0.78}{6} & \scalebox{0.78}{3} & \scalebox{0.78}{2} & \scalebox{0.78}{\secondres{6}} & \scalebox{0.78}{\secondres{10}} & \scalebox{0.78}{2} & \scalebox{0.78}{0} & \scalebox{0.78}{0} & \scalebox{0.78}{0} & \scalebox{0.78}{0} & \scalebox{0.78}{0} & \scalebox{0.78}{0} & \scalebox{0.78}{0} & \scalebox{0.78}{0} & \scalebox{0.78}{0} & \scalebox{0.78}{2} & \scalebox{0.78}{0} & \scalebox{0.78}{0} & \scalebox{0.78}{0} & \scalebox{0.78}{0} & \scalebox{0.78}{0} \\ 
    \bottomrule
  \end{tabular} \label{tablee}
\end{table*} 
\subsection{Ablation Studies}
In this section, we will verify the irreplaceability of implicit regularization. The modules for explicit regularization have been discussed in several classical papers~\cite{guangpu,FreDF}. However, explicit regularization methods have certain drawbacks, such as introducing Fourier noise and significant variations in the amplitudes of frequency components. Many recent studies have highlighted that adaptive normalization methods can alleviate the disadvantages of explicit regularization~\cite{Filter,Freqgui1}. Compared to traditional normalization methods, can the implicit regularization proposed by FreLE better address the shortcomings of explicit regularization? As shown in the ablation and module comparison experiments in Table~\ref{tab:ablationstudy}, the performance of the FreLE module is optimal across the four datasets—ETTm1, ETTm2, ECL, and Weather. It outperforms traditional normalization methods in extracting frequency features from time series.

\begin{table}[H]
    \centering
    \caption{Averaged results for each setting in the ablation study. EFR stands for Explicit Frequency Regularization, IFR stands for Implicit Frequency Regularization, and AN stands for Adaptive Normalization.}
    \resizebox{0.49\textwidth}{!}{
    \begin{tabular}{c|cc|cc|cc}
    \multirow{2}{*}{Setting} &\multicolumn{2}{c}{\textbf{EFR-IFR}}& \multicolumn{2}{c|}{\textbf{EFR}} & \multicolumn{2}{c}{\textbf{EFR-AN}}   \\ 
                             & MSE   & MAE   & MSE    & MAE    & MSE    & MAE    \\ 
                             \hlineB{2.5}
    ETTm1                    & \textbf{0.386} & \textbf{0.394} & 0.411  & 0.432  & 0.407  & 0.435  \\
    ETTm2&\textbf{0.278}&\textbf{0.319}&0.293&0.325&0.280&0.351\\
    ECL&\textbf{0.175}&\textbf{0.271}&0.197&0.311&0.251&0.294\\
    Weather                  & \textbf{0.247} & \textbf{0.277} & 0.254  & 0.291  & 0.255  & 0.283
    \end{tabular}
    }
    \label{tab:ablationstudy}
\end{table}
\subsection{Hyperparameter Sensitvity }
In this section, we conduct a sensitivity analysis on the frequency loss balance hyperparameter. For the ETTm1 and ECL datasets, we select points at 0.1 intervals for $\delta\in[0,1]$ and perform experiments. The relationship between the hyperparameter and model performance is illustrated in Figure A. It can be observed that when $\delta=0$, the model performs worst, as the frequency regularization method is not applied. Additionally, directly setting $\delta=1$ without hyperparameter tuning also yields good experimental performance. This observation is consistent with the frequency decoupling phenomenon discussed in FreDF~\cite{FreDF}. Notably, at $\delta=0.3$, the experimental performance is generally optimal, with the loss values for both frequency domain and time domain losses being nearly identical. This indicates that the best experimental results are achieved when the importance of both tasks is balanced equally.    
\begin{figure}[H]
\centering
    \begin{subfigure}[b]{0.45\textwidth}
        \centering
        \includegraphics[width=1\textwidth]{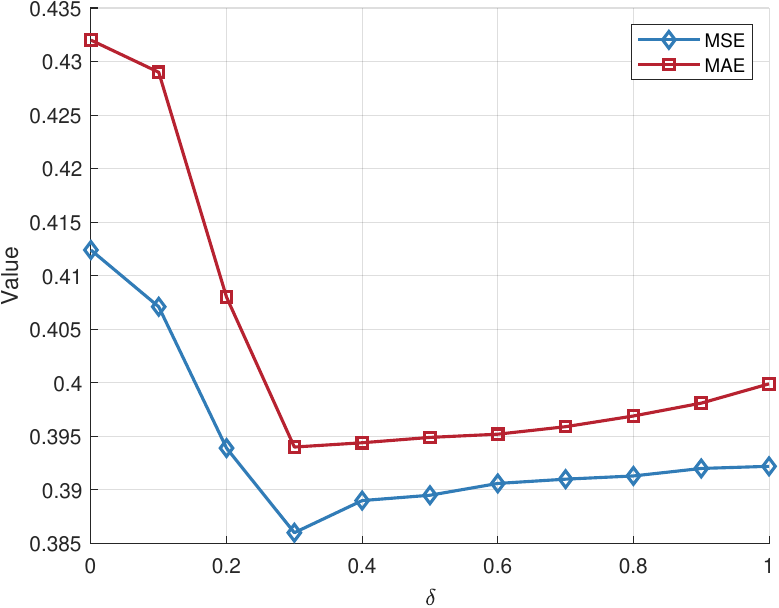}
        \caption{A study on hyperparameter sensitivity based on the ETTm1 dataset}
        \label{fig:subfig1}
    \end{subfigure}
    \hspace{1em}
    \begin{subfigure}[b]{0.45\textwidth}
        \centering
        \includegraphics[width=1\textwidth]{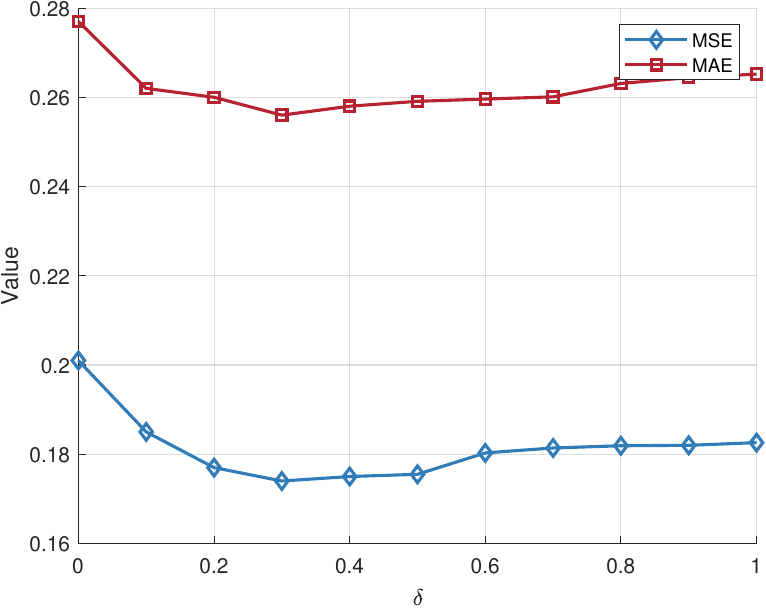}
        \caption{A study on hyperparameter sensitivity based on the ECL dataset}
        \label{fig:subfig2}
    \end{subfigure}
    \caption{Hyperparameter Sensitivity. }
    \label{fig:FLoss-3}
\end{figure}
\subsection{Parameter-Performance Curve Analysis}
\begin{figure}
\centering
    \begin{subfigure}[b]{0.45\textwidth}
        \centering
        \includegraphics[width=1\textwidth]{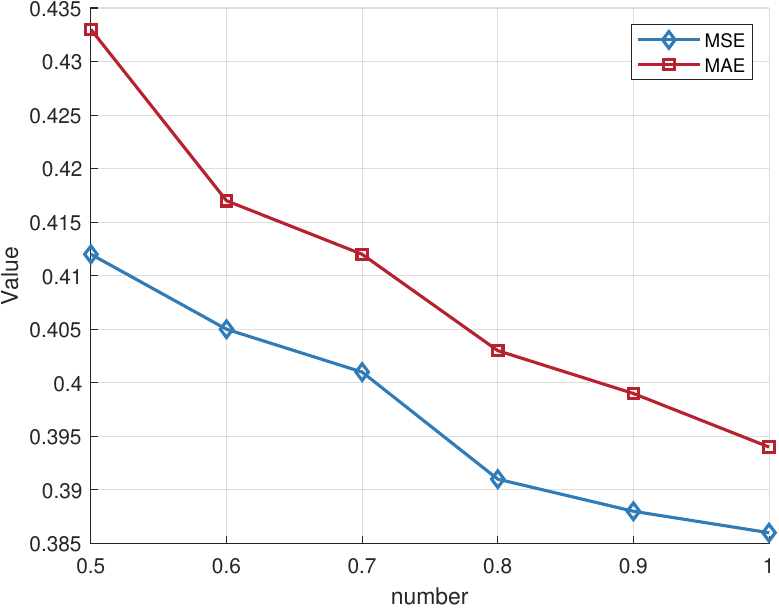}
        \caption{The relationship between the percentage of parameter 'num' retained and model performance on the ETTm1 dataset.}
        \label{fig:subfig1}
    \end{subfigure}
    \hspace{1em}
    \begin{subfigure}[b]{0.45\textwidth}
        \centering
        \includegraphics[width=1\textwidth]{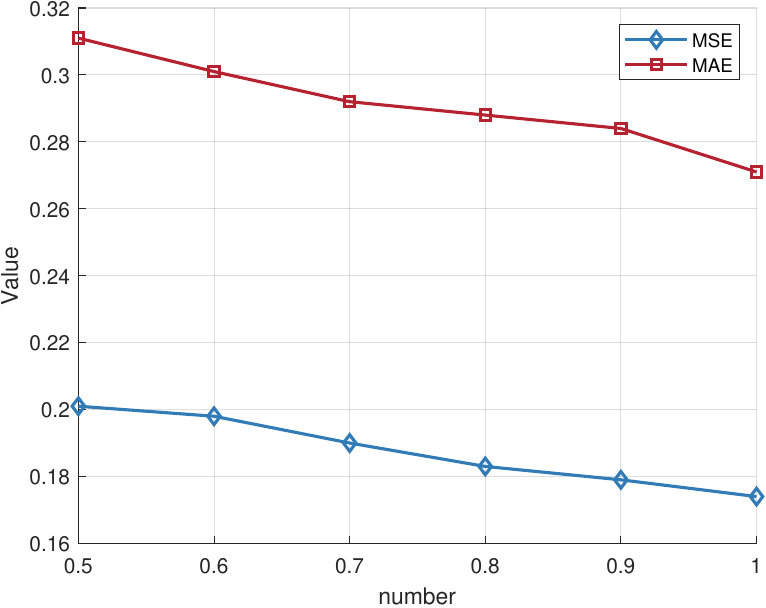}
        \caption{The relationship between the percentage of parameter 'num' retained and model performance on the ECL dataset.}
        \label{fig:subfig2}
    \end{subfigure}
    \caption{Parameter-Performance Curve. }
    \label{fig:FLoss-4}
\end{figure}

In this section, we will reduce the parameter usage of FreLE to analyze its primary impact on the model. The model parameters of FreLE mainly arise from those involved in calculating frequency signals for each layer in explicit regularization. In this section, we will introduce the method of amplitude filtering to reduce the parameter quantity adaptively \cite{LPF-1,HPF-2}: set a threshold $\epsilon_\xi$ for the signal amplitude, and let $\xi_i=0$ if and only if $|\xi|<\epsilon_\xi$. The relationship between parameters and performance is illustrated in Figure~\ref{fig:FLoss-4}, where the number of neural network parameters is defined as $num \in [0.5, 1]$.

When the parameter retention rate is $\text{num}=0.8$, the model has already stabilized, as observed from the experimental results on ETTm1 and ECL.  By reducing 20$\%$ of the model parameters while maintaining performance, the model's performance only decreases by 2$\%$.
\section{Conclusion}
This paper adopts the dynamic approach of spectral bias as its starting point and thoroughly investigates the phenomenon of spectral bias in 2-DNNs and time series models. After validating through extensive empirical experiments that nearly all time series models exhibit the spectral bias phenomenon, we propose the FreLE algorithm, which consists of two modules: explicit regularization and implicit regularization. Our extensive experiments on seven datasets demonstrate the high efficiency of the FreLE algorithm. Furthermore, the ablation experiments, sensitivity analysis, and discussions on computational efficiency indirectly confirm the irreplaceable role of the implicit regularization module in FreLE. In the future, we plan to explore the development of new optimization algorithms by leveraging the implicit regularization method we have adopted, with the goal of better utilizing the information priors provided during the Fourier transformation process to address a broader range of problems.
\section*{Acknowledgement}
This work was supported by the National Natural Science Foundation of China under Grant 62271234, the Open Foundation of State Key Laboratory of Public Big Data (Guizhou University) under Grant No. PBD2022-16, the Fundamental Research Funds for Heilongjiang Universities under Grant 2022-KYYWF-1042, Double First-Class Project for Collaborative Innovation Achievements in Disciplines Construction in Heilongjiang Province under Grant No. LJGXCG2022-054 and LJGXCG2023-028.

\appendix
\section*{Recent Work on Spectral Bias Phenomenon}
Progress in the interpretability of deep learning has been challenging. Compared to traditional modeling theories, the vast number of parameters in deep learning should theoretically suggest a negative outcome: overfitting. However, despite the increasing number of parameters in deep learning network architectures, overfitting, as predicted by traditional modeling theory, does not seem to occur. Thus, developing a robust theoretical understanding of this non-overfitting phenomenon has become increasingly important.

Some researchers aim to establish a theoretical framework for neural networks by beginning with idealized assumptions about DNNs models and applying classical optimization theories through rigorous mathematical proofs. For example, when the width of a neural network approaches infinity, the training dynamics under gradient descent optimization can be approximated by a linearized model governed by the Neural Tangent Kernel (NTK)~\cite{NTK1,NTK2}. Neural networks excel at learning both simple and complex interaction effects within data but struggle with interactions of moderate complexity, a phenomenon known as the "representation bottleneck". 

While these studies provide a solid theoretical foundation for neural networks, they all rely on complex mathematical assumptions. A significant challenge is that during the training process, the gradient's sharpness often exceeds theoretical thresholds~\cite{GD16}, undermining the reliability of some classical assumptions and rendering them insufficient to explain the behavior of general neural networks. Furthermore, a case study suggests that norm-based complexity measures perform poorly in stochastic optimization, sometimes even adversely affecting the generalization of neural networks	~\cite{Shi17}. This reality encourages the exploration of phenomenological approaches to better understand neural network theory.	

The frequency principle is a recently discovered phenomenological approach to explaining neural network phenomena. Xv et al. \cite{Freq1} observed that over-parameterized DNNs tend to use low-frequency functions to fit training data. These networks initially capture the low-frequency components of the training data and while maintaining the high-frequency components at a smaller magnitude. To extract high-frequency components from the training data, techniques such as the discrete Fourier transform or the design of relaxed objective functions can be employed, which convert high-frequency signals to low-frequency signals. The frequency principle has also been applied to guide the solution of partial differential equations (PDEs). Various researchers have repeatedly demonstrated the reliability of this principle. For instance, Rahaman et al.\cite{Rehaman18} proposed the concept of spectral bias in the learning process of neural networks. Additionally, Prateek Verma~\cite{verma19}, in a technical report at Stanford University, introduced the implicit Fourier transform operations within neural network architectures. These studies have thoroughly investigated methods for studying time series in the frequency domain.
Researchers in time series analysis recognized the potential of frequency domain features early on. Work related to frequency domain features has been continuously proposed: as early as Godfrey et al.'s study~\cite{1F20}, Fourier decomposition was used to enhance model generalization. However, early neural network architectureswere unsuitable for multimodal learning in both the time and frequency domains, and significant progress was not made until the introduction of the Transformer deep learning model architecture. 

The self-attention and multi-head attention mechanisms in Transformers significantly accelerated the development of multimodal learning. Since then, many time series forecasting models based on the Transformer architecture that integrate time and frequency domains have been proposed, such as TFT~\cite{Ftran22}, FEDformer~\cite{Fedformer}, and JTFT~\cite{Ftrans23}. The excellent performance of the Transformer architecture in learning frequency domain features has given scholars confidence to apply this method in the MLP domain. Some notable MLP methods, such as Timenet~\cite{Timenet}, FreDF~\cite{FreDF}, and FITS~\cite{FITS},   have been continuously explored by researchers.
\end{document}